\documentclass[sigconf]{acmart}
\AtBeginDocument{%
  \providecommand\BibTeX{{%
    \normalfont B\kern-0.5em{\scshape i\kern-0.25em b}\kern-0.8em\TeX}}}

\setcopyright{acmcopyright}
\copyrightyear{2018}
\acmYear{2018}
\acmDOI{XXXXXXX.XXXXXXX}

\acmConference[Conference acronym 'XX]{Make sure to enter the correct
  conference title from your rights confirmation email}{June 03--05,
  2018}{Woodstock, NY}
\acmBooktitle{Woodstock '18: ACM Symposium on Neural Gaze Detection,
 June 03--05, 2018, Woodstock, NY} 
\acmPrice{15.00}
\acmISBN{978-1-4503-XXXX-X/18/06}

\acmSubmissionID{2831}

\usepackage{balance}       % to better equalize the last page
\usepackage{graphics}      % for EPS, load graphicx instead 
\usepackage[T1]{fontenc}   % for umlauts and other diaeresis
\usepackage{mathptmx}
\usepackage{color}
\usepackage{booktabs}
\usepackage{textcomp}
\usepackage{eurosym}
\usepackage{ljhnick}
\usepackage{mathtools}
\usepackage{multicol}
\newif\ifCOMMENTS
\COMMENTStrue
\ifCOMMENTS

\else

\fi
\begin{document}

\title{\codename{}: A Robotic Data Collection Pipeline for the Pose Estimation
  of Appearance-Changing Objects}


% tentative author list

\author{Jiahao ``Nick'' Li}
\affiliation{%
  \institution{UCLA HCI Research}
  \city{Los Angeles}
  \country{United States}}
\email{ljhnick@g.ucla.edu}
\authornote{Both authors contributed equally to this research.}

\author{Toby Chong}
\affiliation{%
  \institution{TOEI Zukun Research}
  \city{Tokyo}
  \country{Japan}}
\email{tobyclh@gmail.com}
\authornotemark[1]

\author{Zhongyi Zhou}
\affiliation{%
  \institution{University of Tokyo}
  \city{Tokyo}
  \country{Japan}}
\email{zhongyi.zhou.work@gmail.com}

\author{Hironori Yoshida}
\affiliation{%
  \institution{Future University Hakodate}
  \city{Hakodate}
  \country{Japan}}
\email{hyoshida@fun.ac.jp}

\author{Koji Yatani}
\affiliation{%
  \institution{University of Tokyo}
  \city{Tokyo}
  \country{Japan}}
\email{koji@iis-lab.org}

\author{Xiang `Anthony' Chen}
\affiliation{%
  \institution{UCLA HCI Research}
  \city{Los Angeles}
  \country{United States}}
\email{xac@ucla.edu}

\author{Takeo Igarashi}
\affiliation{%
  \institution{University of Tokyo}
  \city{Tokyo}
  \country{Japan}}
\email{takeo@acm.org}


\renewcommand{\shortauthors}{Trovato and Tobin, et al.}


\begin{teaserfigure}
  \centering
    \includegraphics[width=\textwidth]{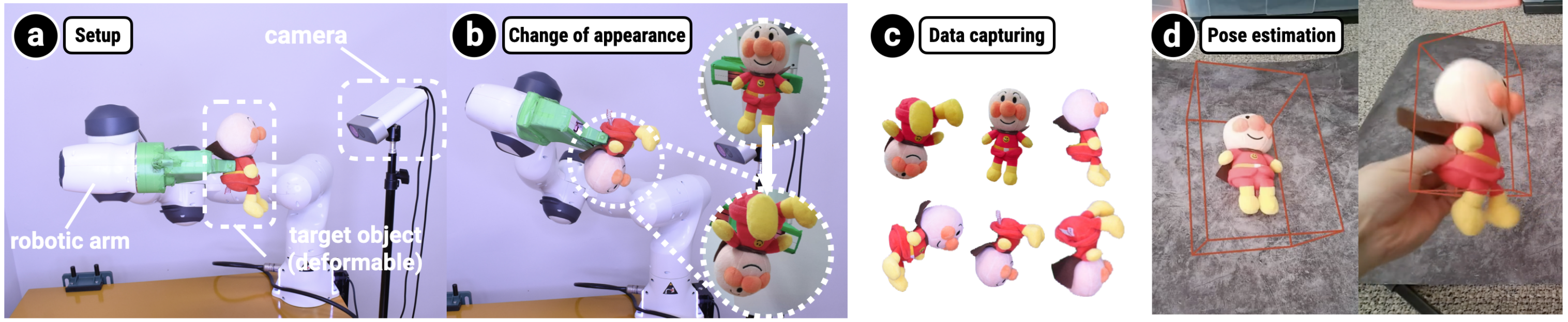}
 \caption{
 % \textit{Workflow of \codename{} pipeline.} 
 The \codename{} pipeline is a robotic system designed to collect datasets for the purpose of pose estimation of appearance-changing objects, \eg a deformable plush toy (a). The system consists of a robotic arm and an RGB camera, which allows for data collection (c) of objects with appearance-changing features (b). Through data augmentation and training on off-the-shelf deep learning models using the collected data, the system can effectively estimate the pose of the plush toy during manipulation, even as it transitions through deformation (d).
 % \eg a clamp actuable with multiple states (open v.s closed).
 % The system consists of a robotic arm and an RGB camera, which allows for data collection of objects with appearance-changing feature such as the clamp with multiple states (a).
 % \codename{} is capable to capture the data in different mechanical states of the clamp (b) in various 6D configurations (c). 
 % Through data augmentation and training on off-the-shelf deep learning models using the collected data, the system can effectively estimate the pose of the clamp during manipulation, even as it transitions through different states (d).
 % transparent objects like the transparent flask (top) and mechanically state-changing objects like the clamp (bottom). The system can collect RGB images of the target objects in 3652 poses (b) and can automatically annotate the data with ground truth pose relative to the camera (c). Through data augmentation and training on off-the-shelf deep learning models, the \codename{} pipeline is capable of estimating the pose of appearance-changing objects during interaction (d).
 }
 \label{fig:teaser}
\end{teaserfigure}

\begin{abstract}

Object pose estimation plays a vital role in mixed reality interactions when user manipulate tangible objects as controllers. 
Traditional vision-based object pose estimation methods leverage 3D reconstruction to synthesize training data. 
However, these methods are designed for static objects with diffuse colors and do not work well for objects that change their appearance during manipulation, such as deformable objects like plush toys, transparent objects like chemical flasks, reflective objects like metal pitcher, and articulated objects like scissors. 
To address this limitation, we propose \codename{}, a robotic pipeline that emulates human manipulation of target objects while generating data labeled with ground truth pose information. The user first gives the target object to a robotic arm, and the system captures many pictures of the object in various 6D configurations. The system trains a model by using captured images 
% \xac{why pair?} 
and their ground truth pose information automatically calculated from the joint angles of the robotic arm.
We showcase pose estimation
% the potential of achieving pose estimation
for appearance-changing objects by training simple deep-learning models using the collected data and comparing the results with a model trained with synthetic data based on 3D reconstruction via quantitative and qualitative evaluation. The findings underscore the promising capabilities of  \codename{}.

  % \xac{overall: to make the system more relevant to uist, maybe we can frame it as `RoCap: robotic tool support for collecting ...'}
\end{abstract}





\maketitle

\section{Introduction}

% \missingfigure[]{teaser figure}
% background

% Interacting with tangible objects significantly enhances the mixed reality experience, particularly for applications such as storytelling, skill training, and education. 
% % Beyond traditional methods involving controllers (\xx) or specifically designed proxies (\xx), employing existing tangible objects could create a more immersive and intuitive experience.

% Employing existing tangible objects as controllers could provide a more immersive experience (\eg story telling using existing plush toy) or diversed tangible motions(\eg various handheld tools)

% Employing existing tangible objects could 

Leveraging existing tangible objects as controllers in mixed reality (MR) can significantly enhance the immersive experience and allow for a wider range of tangible motions, particularly for applications such as storytelling, skill training, and education. 
For example, employing plush toys to guide storytelling allows for a personalized and engaging experience, and using various handheld tools can facilitate more precise and diverse interaction mechanisms to make tasks feel natural and intuitive.
% plush toys can be used to guide an immersive storytelling and various handheld tools can be used to enable more precise and varied interaction mechanisms. 
Such an approach demands the capabilities for accurate predictions of 6D pose estimation --- identifying an object's location and orientation in the 3D space.

% Accurately predicting the 6D pose (position and orientation) of a physical object is a crucial task for a wide range of applications in the field of robotics and augmented reality (AR) where robots or human users need to interact with these objects in the environment. 
Vision-based pose estimation has gained popularity in the past few years over tracker or sensor based methods, as it does not require additional hardware, alter the appearance or interfere with the normal use of the objects and it is cost effective and accessible. 
% However, estimating the 6D pose of an object from an unconstrained RGB 
% % \xac{depth camera is also common---why not comparing with it? (and why not using depth cam in our system?)} 
% image remains highly challenging due to the ambiguity nature of the estimation.
% To solve this problem, 
Researchers have adopted different approaches including mapping image feature to the 3D model of the object \cite{hettiarachchi2016annexing} and matching point cloud constructed by depth camera \cite{barnes2008video}. 
More recently, data-driven deep learning methods~\cite{xiang2017posecnn, peng2019pvnet} demonstrated accurate predictions of the 6D pose of pre-defined sets of object included in carefully crafted datasets \cite{calli2015ycb, krull2015learning}.
However, it remains unclear how well they work on objects where carefully labeled data do not exist such as personal objects.
To address this issue, some prior work enables end-users to collect datasets for everyday objects 6D pose estimation \cite{marion2018label, qian2022arnnotate}, introducing synthetic approaches to generate a large amount synthetic data given the 3D model of the objects \cite{dwibedi2017cut}, or adopts a few-shot learning method by training on 3D mesh reconstructed from a short clip of video \cite{liu2022gen6d}.

% The former mainly focuses on rigid and static object
A limitation of these existing methods is that they mainly focus on objects that are static objects with diffuse colors, with a less focus on objects that \textbf{change their appearances} when being manipulated, including objects with challenging appearance materials (\eg transparent and specular objects), deformable objects and articulated objects~\cite{thalhammer2023challenges}. 
% For example, a plush toy my dramatically change its physical appearance due to the manipulation and a model trained on the images of a static plush toy might produce lower accuracy at recognizing the same toy deformed due to manipulation and gravity.
A pair of scissors will dramatically change its physical appearance due to mechanical operation and a model trained on the image of a closed pair of scissors might produce lower accuracy at recognizing the same pair in an open configuration. 
% Similarly, a pair of scissors will have multiple mechanical
Similarly, a plush toy that changes its shape during manipulation when being affected by gravity will affect the performance of the pose estimation.
While one intuitive approach is to capture data while a human user is manipulating the objects, annotating such data at scale would be costly and error-prone.

% while bridging the gap that human users are not able to label the ground truth of the 6D pose \cite{marion2018label, qian2022arnnotate}.
% These work focuses on objects that are rigid and static or being held by human hand, with a less focus on objects that will \textbf{change their states} during manipulation.
% For example, a pair of scissors will dramatically change its form due to mechanical operation and a model trained on the image data in which the scissors close will fail when the pair of scissors is in the opening configuration. 
% Or a soft toy will change its appearance during manipulation when being affected by gravity will affect the performance of the pose estimation.
% While one intuitive approach is to capture data while a human user is manipulating the objects, the 6D pose will then be extremely hard to label in this configuration\zhongyi{}{ without any support from markers}.

% \nick{add the reason why using synthetic data is not enough}

To address the challenge, we propose \codename{}, an automated pipeline to collect image data of appearance-changing object for 6D pose estimation using a robotic arm with minimum human intervention. 
% \nick{need to add some of the motivation here: robotic arm is not as accessible as mobile phone, how do we frame it?}
% Although robotic arms may not be as accessible as mobile phones, they offer precise control and the ability to replicate complex manipulation tasks, making them well-suited for this application. In the context of mass manufacturing or future sharable robotic arms, \codename{} could serve as a valuable tool for streamlining processes and enhancing automation capabilities.
We deploy a robot arm to mimic human's hand to manipulate the objects while capturing the image data as shown in \fgref{teaser}.
The 6D poses of the object of each image can be obtained with robotic forward kinematics as each joint of the robotic arm is precisely controlled. Specifically, \codename{} performs the data capturing process for eight different appearance-changing objects with deformable, transparent, reflective and articulated properties (Figure \ref{fig:example_objects}).  
% \xac{eight different kinds of objects? which eight?}

We also implemented a simple pose estimation pipeline to quantitatively and qualitatively evaluate the pose estimation performance of the model trained on our collected data comparing against a few-shot learning pose estimation approach based on 3D reconstruction (Gen6D~\cite{liu2022gen6d}).
Both the quantitative and qualitative evaluation results demonstrate that existing work struggles with appearance-changing objects and our approach shows promise in overcoming these limitations with improved pose estimation accuracy.
% is limited in address appearance-changing objects and our method shows potential in addressing the limitation by improving the pose estimation accuracy.
% our method can effectively address some of the limitations of existing work, and seamlessly integrate with current 6D pose estimation frameworks to advance research in the field. 
% \nick{might not need to use the result of crowd workers, or we can put them into the appendix}

In summary, our contributions are two-fold:
\begin{itemize}
    \item \textbf{A robotic data collection pipeline} with a 6 DoF robotic arm which captures and annotates 6D pose data for objects that change their appearance during manipulation, addressing limitations in existing data collection methods.
    \item \textbf{Quantitative and qualitative evaluations} to demonstrate the feasibility of the pipeline via improved accuracy of appearance-changing objects pose estimation by comparing with an advanced pose estimation method in the field of computer vision.
    % the feasibility of the pipeline by training existing models on the collected data and comparing the performance with state-of-the-art pose estimation methods to demonstrate its potential to advance research in the field.
    % \item \textbf{Datasets of eight objects with labeled ground truth \xx}
    % the data collected to train on existing deep learning framework to showcase applications.
\end{itemize}

\section{Related Work}

\begin{figure}[t]
    \centering
    \includegraphics[width=0.9\columnwidth]{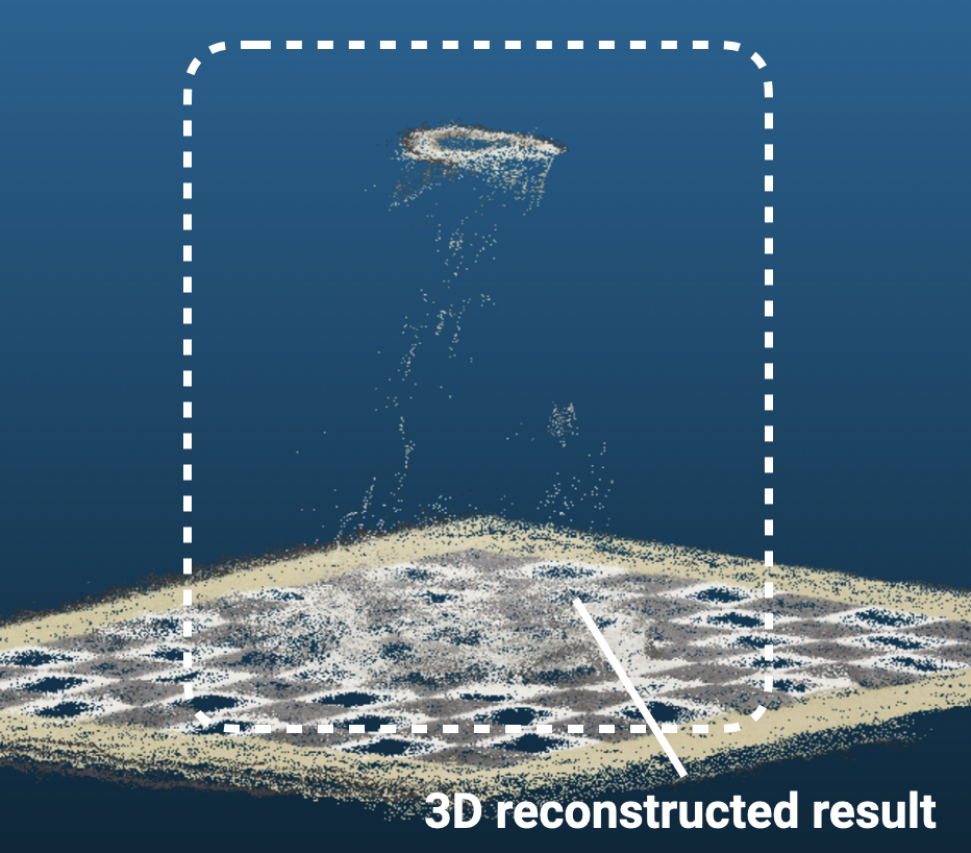}
    \caption{3D reconstructed results for a transparent flask.}
    \label{fig:3d}
\end{figure}

\subsection{Object pose estimation}
Object pose estimation plays a crucial role in various HCI applications such as augmented reality \cite{suzuki2020RealitySketch, hettiarachchi2016annexing, barnes2008video, held2012puppetry} and robotics and automation \cite{li2022roman}. Over recent decades, researchers have explored diverse approaches to predict an object's pose. 
These includes sensor applications like IMUs, physical marker techniques such as fiducial markers \cite{kalaitzakis2021fiducial, ulrich2022towards, garrido2014automatic}, optic trackers \cite{zhou2020gripmarks} 3D printed embedded QR code \cite{dogan2022infraredtags}, computer vision techniques such as color-based tracking \cite{suzuki2020RealitySketch}, feature point tracking \cite{barnes2008video} and point cloud alignment \cite{hettiarachchi2016annexing}.
Recent advancement in deep learning has unlocked new challenging tasks such as predicting the poses of hand-object interaction \cite{hampali2020honnotate,liu2021semi,chao2021dexycb}, articulated objects \cite{li2020category} and other problem setups \cite{lugaresi2019mediapipe, Ahmadyan2021objectron, rukhovich2021imvoxelnet, yang2022boosting}.
Extending this line of research, \codename{} focuses on a new problem setup where the objects will change their appearance during the manipulation.
Note that \codename{} does not contribute new model architecture or algorithm to improve the performance in the field of deep learning. Instead, \codename{} contributes a novel data collection method and 
the data captured by the system can serve as great resources for researchers in the community of computer vision and machine learning to solve the downstream tracking problems.

% This ranges from applying sensors such as IMU, to physical marker based approaches \cite{kalaitzakis2021fiducial, ulrich2022towards, garrido2014automatic, zhou2020gripmarks, dogan2022infraredtags}, to computer vision based approach such as color-based tracking \cite{suzuki2020RealitySketch}, feature points tracking \cite{barnes2008video} or points cloud alignment \cite{hettiarachchi2016annexing}, to more recently, data-driven deep learning approaches in different task settings such as hand-object interaction \cite{hampali2020honnotate,liu2021semi,chao2021dexycb}, articulated objects \cite{li2020category} and other problem setups \cite{lugaresi2019mediapipe, Ahmadyan2021objectron, rukhovich2021imvoxelnet, yang2022boosting}. 

 \begin{figure}
    \centering
    \includegraphics[width=1\columnwidth]{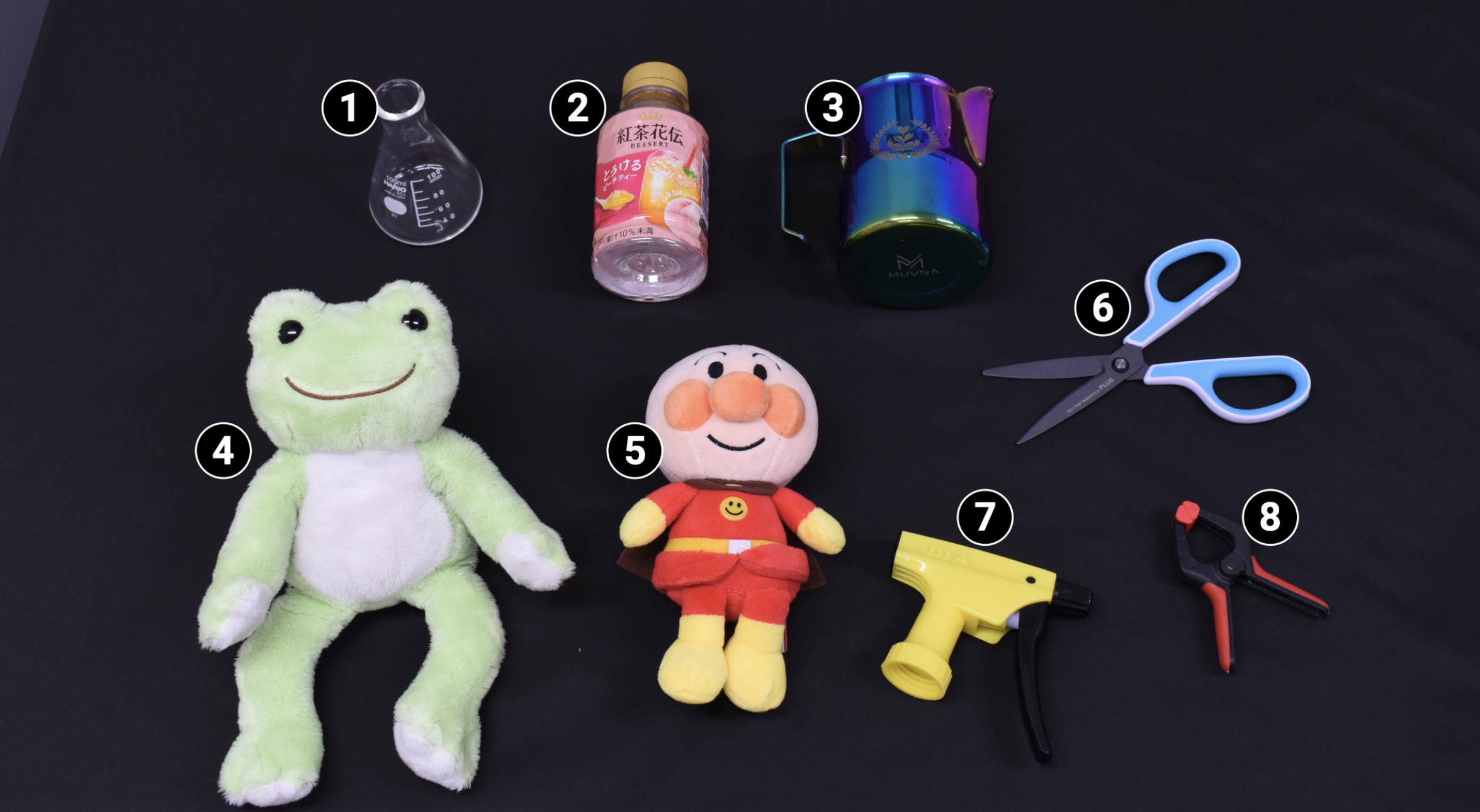}
    \caption{Example objects for each category that \codename{} is focusing on, \textbf{\textit{Viewing-angle dependent}}: (1) flask, (2) water bottle and, (3) pitcher, \textbf{\textit{Deformable}}: (4) flexible frog and (5) stiff anpanman, 
    \textbf{\textit{Articulated}}: (6) scissors, (7) spray head and (8) clamp.}
    \label{fig:example_objects}
\end{figure}

\subsection{Pose data collection}
Data-driven deep learning approaches require data annotated with ground truth labels. Yet, annotating 6D pose data is challenging, as it is hard to specify 3D bounding box on a 2D image. 
To address this, researchers have investigated various methods including three primary strategies:
\one training on synthesized data, \two utilizing publicly available datasets and \three designing interactive tools for data collection.
% Researchers have explored various approaches to address this challenge, in summary, there are three major approaches in the pose estimation field \one synthesized data, \two publicly available dataset and \three interactive tools to collect data.

\subsubsection{Synthetic data}
One typical way is synthesizing data with the available resources such as the 3D model of the objects. 
This approach is commonly used in tasks such as object segmentation \cite{ros2016synthia} and object detection \cite{dwibedi2017cut}.
And a standard way of using synthetic data in pose estimation is to obtain the 3D model and texture of the objects first and then render them with different target background \cite{su2021synpo}. 
Although synthetic data can be easily scaled, it comes with the drawback of a disparity between real and virtual data, which might impact model performance. Moreover, as illustrated in Figure \ref{fig:3d}, the necessary step of object reconstruction may fail for our target objects, such as the flask.

% With this approach, large enough data can be generated by controling the sampling rate of different poses to render.
% However, it also comes with the limitation that there is a gap between the real data and virtual data which may affect the performance of the model
% % \zhongyi{cause the performance of the model to be not ideal}{harm the performance of the model}. 
% Some work address this by combining both the real and virtual data \cite{xiang2017posecnn}, but
% most importantly, this approach does not apply to our target objects because the objects changes their appearance during manipulation. 
% Although the recent work by Sin \etal trained on synthesized data of deformable objects by simulating the cage-based deformation \cite{sin2022tracking}, it does not apply to other types such as objects with multiple mechanical states.

\subsubsection{Real-world data}
An intuitive way to bypass the issue of synthesized data is to collect data in the real world. 
In the recent years, researchers have adopted two major types of data collection methods. 
The first is \textbf{``static object + moving camera''}, where the pose of the object is calculated from the pose of the camera, which can be read from the embedded sensor. Normally it requires a certain level of human labor as first couple frames need to be manually annotated by matching the 3D model to the physical object. For example, several publically available datasets have been collected in this way for benchmarking in the pose estimation domain, such as YCB Video dataset \cite{xiang2017posecnn}, Linemod \cite{hinterstoisser2013model, brachmann2014learning} and T-Less \cite{hodan2017t}. Additionally, researchers have also developed interactive data collection pipeline to collect data on custom objects (\eg Label Fusion~\cite{marion2018label}. However, since the objects remain static, it is challenging to capture the appearance-changing features.

Another approach is \textbf{``moving objects + static or moving camera''}. While effective for capturing appearance-changing objects, this approach poses challenging for labeling ground truth.
% This approach is suitable for capturing data of appearance-changing objects, however, is challenging in labeling the ground truth data. 
For instance, ARnnotate, used in augmented reality~\cite{qian2022arnnotate}, requires users to hold and move the object along a recorded path, leading to potential errors, especially with objects like articulated items or deformed plush toys.
\codename{} adopts this approach and ensures the labeled ground truth to be precise by calculating the robotic arm's forward kinematics while it manipulates the object to capture the appearance-changing features.

\section{Appearance-changing objects}
In this section we define and explain the importance of three categories of appearance-changing objects that we aim to track using \codename{}. We collected and captured eight items from the three categories with \codename{}. 

\subsection{Deformation}
Deformation refers to changes in the shape or size of an object due to external forces applied during manipulation (i.e., force of the hand and gravity). 
Objects with naturally deformable features can include soft and malleable objects such as fabric materials, clothing and plush toys/stuffed animals.
During manipulation, the objects are affected by gravity all the time, leading to the deformation while the user is moving the objects into different orientation. 
We picked two plush toys of different stiffness, anpanman (stiffer) (Figure \ref{fig:example_objects}(5)) and frog (more flexible)(Figure \ref{fig:example_objects}(4)) as examples of the deformable objects. 
% The pose estimation result is demonstrated in Figure \ref{fig:results}a.
% % While the body of the anpanman is rather rigid, the feet and the arms of it are very soft and easily deformed in different configuration (Figure \ref{fig:appearance}a). 
% Models trained on the static state of the anpanman may struggle when the user is holding the object upside down (\eg in a story telling examples \cite{held2012puppetry} or using this as an input controller \cite{kawasaki2005vision}).

% to demonstrate the challenge of pose estimation for objects 

% As is shown in Figure \xx, \nick{describe the result of tracking the anpanman using gen6d}
% \nick{and describe our result tracking the anpanman}

% may undergo stretching, compression, bending, or twisting, leading to changes in their shape and size. Accurate pose estimation for these objects can be challenging because their appearance can change significantly as they deform, making it difficult for the model to recognize the object and track its pose.

\subsection{Viewing-angle dependent}
The visual appearance of viewing-angle dependent objects includes two main sub-categories of objects, transparent objects (e.g., glass) and reflective objects (e.g., polished metal). 
Appearance of transparent objects depends on the background behind them, which may contain the environment and the user's hands. Tracking and estimating the pose of such transparent objects is a known challenge \cite{fan2021transparent} and hand manipulation may make this even harder.
Appearance of reflective objects on the other hand depends on the environment in front and around it. 
We picked a conical flask and a plastic bottle(Figure \ref{fig:example_objects}(1, 2)) as representations of transparent objects of different level of translucency (Figure \ref{fig:appearance}b). We also included a reflective pitcher to represent reflective object. 
% Existing method does not support tracking such transparent objects as the 3D reconstruction would fail (Figure \ref{fig:3d}).
% Figure \ref{fig:results}b shows the pose estimation results.
% As is shown in Figure \xx, \nick{describe the failure case of using the gen6d on transparent objects}
% However, with the model trained on our data, we are able to achieve a viable pose estimation of the objects \nick{provide more details}

% \missingfigure[]{examples of the flask}

% \nick{if we do not have bandwidth, we will keep it that way. or we can add a few more objects such as the reflective one}

% \paragraph{Appearance}
% \begin{itemize}
%     \item transparent
%     \item due to reflection
% \end{itemize}

% \paragraph{Shape}

% \begin{itemize}
%     \item deform due to gravity
% \end{itemize}

\subsection{Articulated}
Objects with articulated features refer to objects whose appearance changes through manual manipulation or interaction. 
These changes can occur due to the inherent function of the physical objects. For examples, various handheld tools will change their mechanical forms while being manipulated by human.
% These objects can include tools, toys, and devices that are designed to be interacted with by humans to change their state or function.
% Human may also intentionally change the state of the object in different applications. For example, the user may change the shape or visual appearance of a puppetry in a storytelling scenario \cite{sin2022tracking}.
We selected three manually-changing objects: a clamp, a pair of scissors and, a head of spray bottle to represent two different types of manual gripping and hand operation (holding and pinching). 

\begin{figure*}
    \centering
    \includegraphics[width=\linewidth]{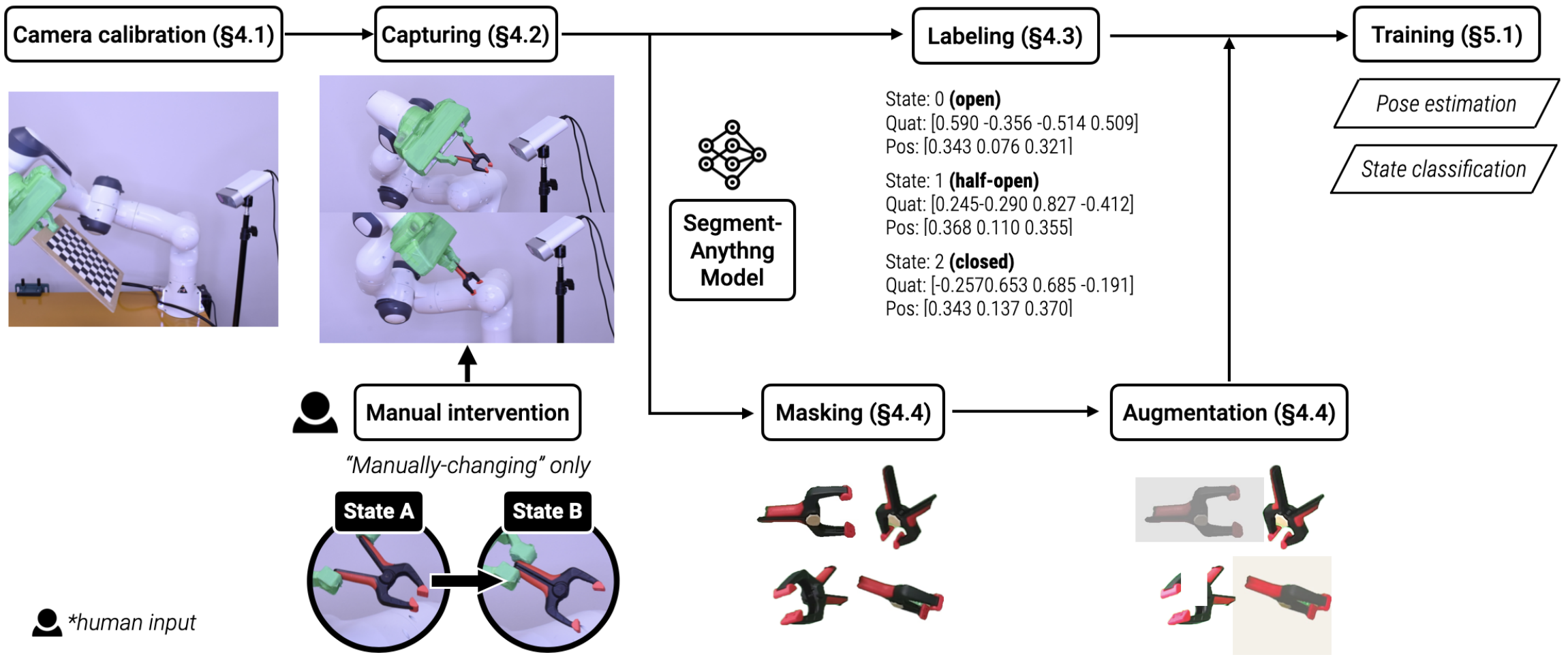}
    \caption{\textit{Overview of \codename{}.} \codename{} pipeline consists of camera calibration (\S\ref{sec:calibration}), data capturing (\S\ref{sec:data_collection}), data labeling (\S\ref{sec:labeling}),  data processing (\S\ref{sec:processing}) and data augmentation (\S\ref{sec:processing}). By training on an existing deep learning framework, \codename{} achieves object segmentation, state classification and pose estimation for appearance-changing objects.}
    \label{fig:workflow}
\end{figure*}

\section{\codename{} pipeline}

% \nick{discuss the overall research question first, and there are several specific research questions for using robotic arm to answer. Simple Copy-Paste is a Strong Data Augmentation Method
% for Instance Segmentation. cite this paper for the data augmentation process}

In this section, we will introduce the design of the \codename{} pipeline, which is easily replicable using any 6-DoF robotic arm, we document the essential knowledge and technical challenges addressed including \one camera calibration, \two data collection, \three data labeling and \four data pre-processing. 
Figure~\ref{fig:workflow} shows the overview of the \codename{} pipeline and we discuss each step in details as follows.

\subsection{Eye-to-hand camera calibration}
\label{sec:calibration}

\begin{figure*}[t]
    \centering
    \includegraphics[width=\linewidth]{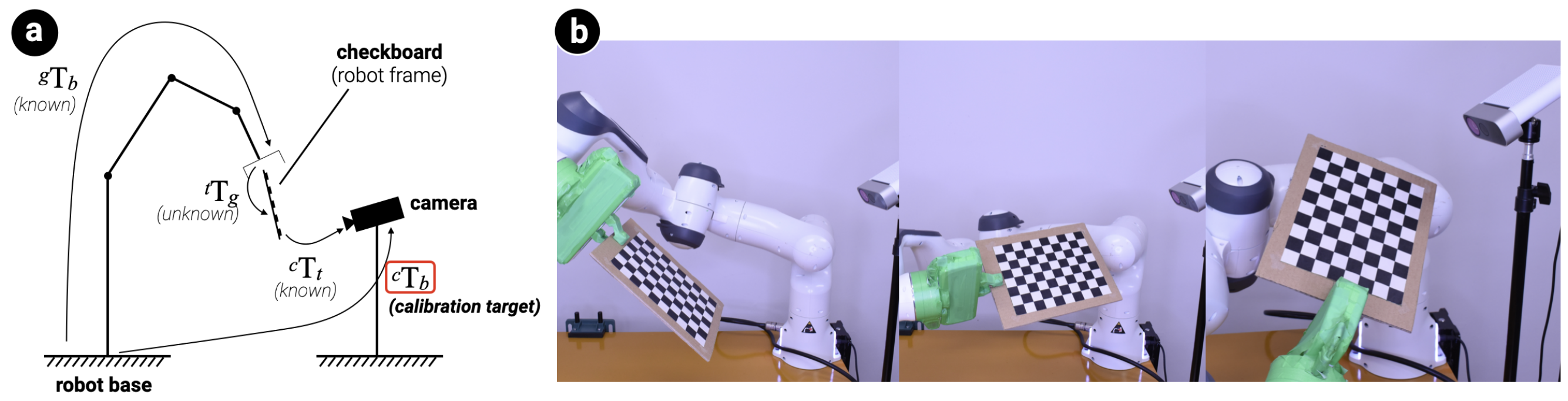}
    \caption{Illustration of the eye-to-hand camera calibration (a). The robotic arm grip a checkerboard and move to multiple positions and orientations for an accurate calibration (b).}
    \label{fig:calibration}
\end{figure*}

The first step of \codename{} pipeline is to calibrate the camera to the robotic arm (Figure \ref{fig:workflow}a).
% As is previously shown in Figure \ref{fig:teaser}a, d
During data collection, the robotic arm will hold the target object using a gripper and the camera is standing on the side to capture the images.
In this setup, the pose of the object in the image refers to the homogeneous transformation of the object from its reference frame to the camera's reference frame.
This is a typical hand-eye calibration problem because as shown in Figure \ref{fig:calibration}.
Assuming the object has the same pose as the end-effector, the goal is to calculate the transformation matrix of the gripper to the camera: $^{c}T_{g}$, which can be calculated from the following equation:
\begin{equation}
\label{eq:equation}
    ^cT_g \ = \ ^cT_b \cdot ^bT_g
\end{equation}
In which $^bT_g$ refers to the transformation from the gripper to the base of the robotic arm which could be calculated by forward kinematics \cite{denavit1955kinematic} while $^cT_b$ refers to the transformation from the base of the robotic arm to the camera frame, which is unknown.

To calculate $^cT_b$, a camera calibration step is required which can be accomplished by using a checkerboard with known size of the squares, which is illustrated in Figure \ref{fig:calibration}b .
% \nick{have not added to the figures yet}. 
By moving the robotic arm to multiple configuration, $^cT_g$ can be calculated from the following $AX = XB$ equations:

\begin{equation}
\begin{gathered}
    ^gT_b^{\left(1\right)} \ ^bT_c \ ^cT_t^{\left(1\right)} = ^gT_b^{\left(2\right)} \ ^bT_c \ ^cT_t^{\left(2\right)} \\
    \bigl({^gT_b^{\left(2\right)}}\bigr)^{-1} \ ^bT_g^{\left(1\right)} \ ^gT_c = ^gT_c \ ^cT_t^{\left(2\right)}\bigl(^cT_t^{\left(1\right)}\bigr)^{-1} \\
    A_i X = X B_i
\end{gathered}
\end{equation}
Here $^cT_t$ refers to the transformation from the checkerboard to the camera frame, which can be calculated knowing the size of the pattern \cite{park1994robot}.
Then the calibration target $^cT_b$ can be calculated from Eq. \ref{eq:equation}.

After the camera is calibrated, the next step is to capture the image data of the objects.

% 1. pipeline design

\subsection{Data collection}
\label{sec:data_collection}
As mentioned in the previous sections, \codename{} collect data of the objects that exhibit the appearance-changing features. 
% More specifically, \codename{} collects state-changing objects in two different categories: 1) objects with naturally-changing features and 2) objects with manually-changing features. We will have a detailed discussion on the categorizations and the choice of examples in each category in Sec \ref{sec:objects}.
% Here we will showcase one example of each category to demonstrate the process of using \codename{} to collect data for object pose estimation.
% 
More specifically, \codename{} collects objects categorized in four types of appearance-changing features: deformable, reflective, transparent and articulated. 
% \missingfigure[]{a figure shows the coverage of the poses in our system and the state changing during the capturing}

\begin{figure}
    \centering
    \includegraphics[width=0.9\columnwidth]{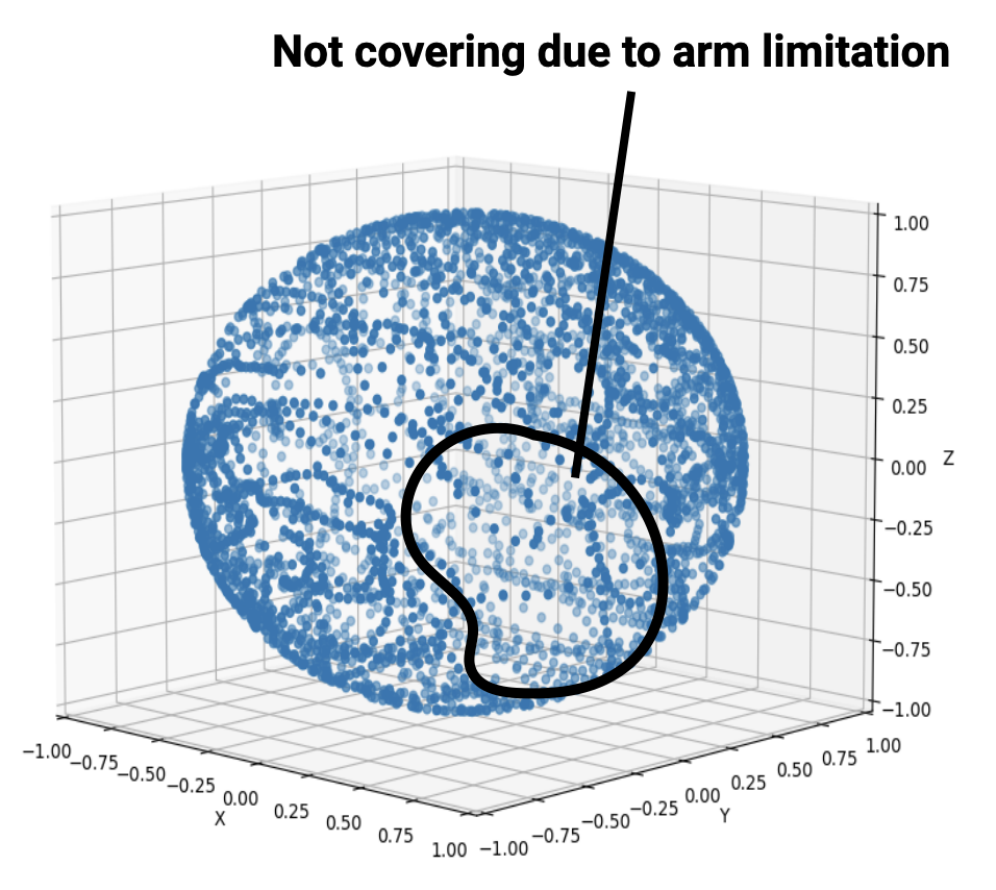}
    \caption{Pose coverage in \codename{} capturing pipeline.}
    \label{fig:pose}
\end{figure}

\subsubsection{Pose coverage}
The goal of the capturing is simple: capture the images of the objects from as many angles as possible to have a good coverage of all the potential pose.
% 
% 
% Quaternions can represent each rotation without causing any ambiguity, however directly sampling quaternions is challenging. Therefore we achieve the coverage of poses by sampling the Euler angles with a specific step of degrees for each yaw, pitch, and roll channel . Subsequently, these Euler angles are converted into quaternions, which are then used to calculate the arc distance between each orientation. This approach is taken because sampling Euler angles can result in significant redundancies. By calculating the arc distance of quaternions, we can eliminate such redundancies, although directly sampling quaternions is challenging.
Quaternions possess the advantage of representing each rotation without introducing any ambiguity. However, directly sampling quaternions proves to be a challenging task. To overcome this obstacle and achieve comprehensive coverage of poses, we opt for sampling Euler angles with a specific step of degrees for each yaw, pitch, and roll channel.
Once we have obtained the Euler angles, they are converted into quaternions. These quaternions are then utilized to calculate the arc distance between each orientation. This methodology is employed due to the inherent redundancies that can arise from sampling Euler angles. By computing the arc distance of quaternions, we effectively eliminate these redundancies. The threshold for eliminating redundancies is set at 0.35, roughly equivalent to a 20\degree{} azimuth angle.

However, due to the hardware limitation of the robotic arm (\eg the joints may have a limited range of motion), \codename{} cannot cover the whole possible poses sampled in this process. 
We use the inverse kinematics solver and path planners in ROS and achieve the final sampling of the poses \codename{} supports.
Figure \ref{fig:pose} visualizes the coverage of the poses in \codename{}. 
% \nick{if possible, also show the comparison with the covered poses of existing method} 
Noted that in existing data collection method where the objects are placed on floor, there will be at least half of the poses not capturable because it is occluded by the contacting ground. 

\begin{figure*}
    \centering
    \includegraphics[width=\linewidth]{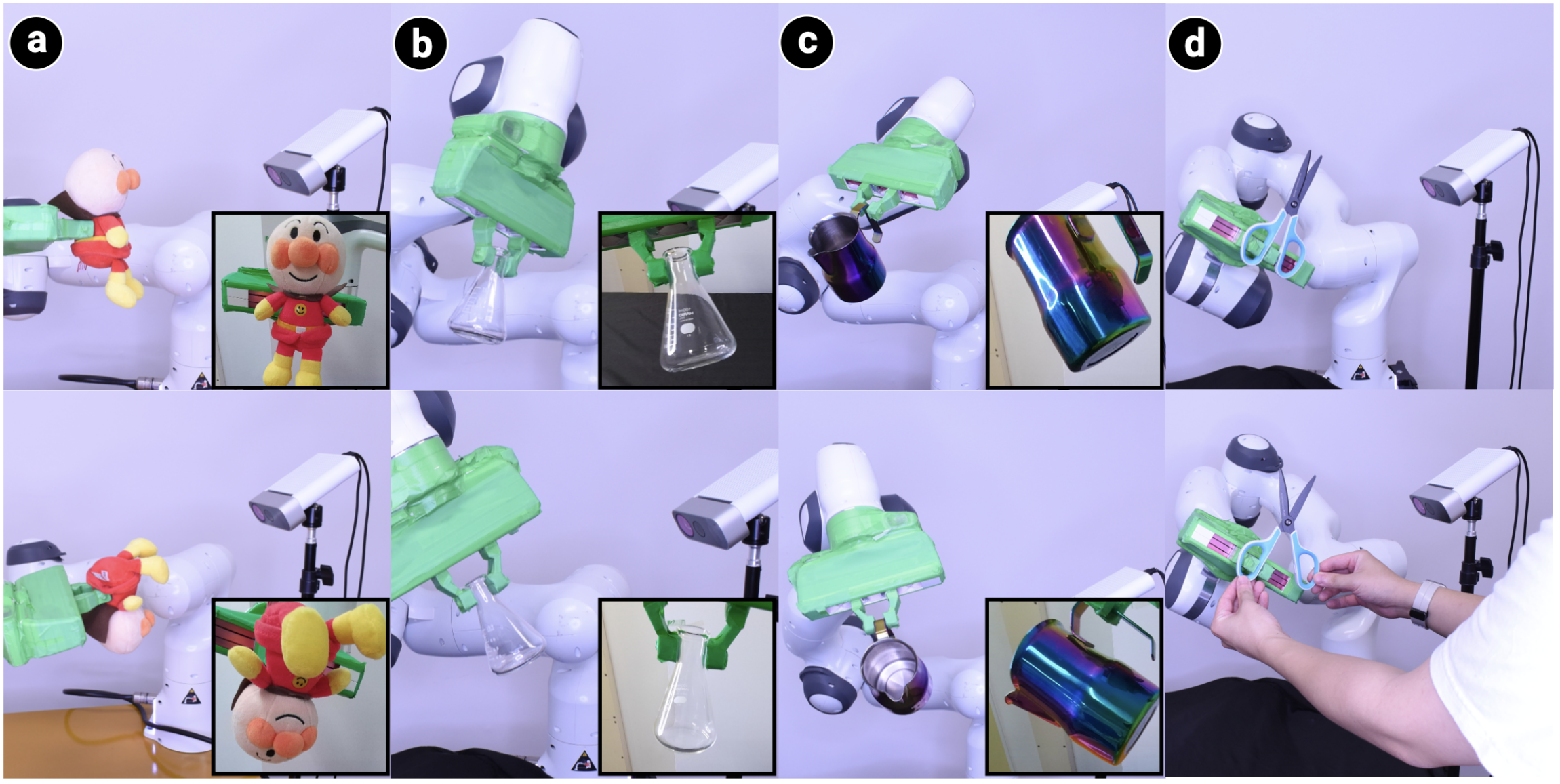}
    \caption{RoCap captures the appearance-changing feature of deformable objects (a), viewing-angle dependent objects including transparent objects (b) and reflective objects (c), and objects with articulated features (d). Human operator is needed if the robotic arm is not able to change the states automatically (d).}
    \label{fig:appearance}
\end{figure*}

\subsubsection{Capturing process}
During the capturing, a human user will be required to hand the target object to the robotic arm and the robotic arm will move along the designed path and the camera capture the RGB images on each sampled point.

For deformable, reflective and transparent objects, there is no further actions from the users as the change of the appearance happens naturally when the object is oriented to different direction while being manipulated by the robotic arm \ref{fig:appearance}abc.
For articulated objects, actions need to be taken in order to change the mechanical states of the objects. The manually-changing action can be achieved either by human or the robotic arm automatically depending on the capability of the robotic arm to change the appearance.
As is shown in Figure \ref{fig:teaser}b, the size of the clamp is small enough to be grasped by the gripper. And the clamp is expected to have multiple states such as closed, open, and mid-open states. Without the help of human, the gripper could be able to change the states of the clamp by applying different forces on the parallel grippers. 
However, for the pair shown in Figure \ref{fig:appearance}d, the robotic gripper is not able to automatically change the states because the handle is too wide for the parallel gripper when it is in open state. 
This is a typical robotic manipulability problem as mentioned in \cite{li2022roman}. 
% the scissors do not have restoring force so that the gripper cannot apply a certain to hold the scissors in an opening state. 
For the case that a robotic arm cannot establish firm gripping on the object, a human operator will be required to manually change the opening angle of the scissors in the interval between the capturing of different states.

\subsection{Data labeling}
\label{sec:labeling}
The transformation from the base frame of the robotic arm to the target object is logged for each captured image in a $4 \times 4$ homogeneous transformation matrix. 
Then the transformation from camera frame to the object could be calculated using Equation \ref{eq:equation}. The rotation and the translation serve as the 6D pose label for the object in each image as shown in Figure \ref{fig:workflow}c left.
% \nick{somewhere to mention we do not need 3D model in previous sections}

\begin{figure*}
    \centering
    \includegraphics[width=\linewidth]{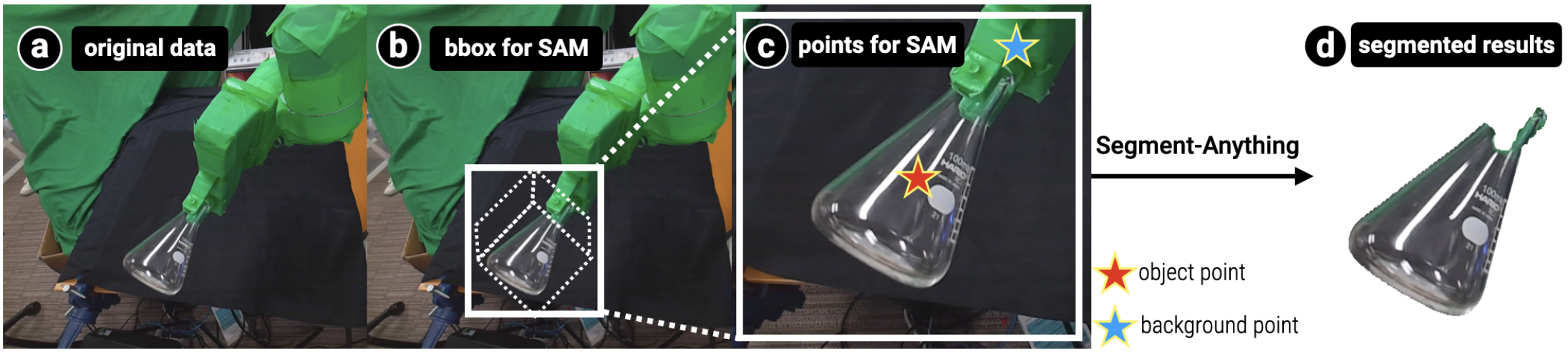}
    \caption{Data processing of data collected in by \codename{}. \codename{} generates mask for each image (d) by prompting SAM with bounding box (b) and points (c). 
    % crops the data and adopts background subtraction (c) of original data (a) and background (b) and green background removal to mask the object (de). \nick{to change}
    }
    \label{fig:processing}
\end{figure*}

\subsection{Data processing and augmentation}
\label{sec:processing}
After capturing the data with the ground truth labels of objects using \codename{}, crucial processing step must be performed to facilitate subsequent pose estimation training. 
A typical object pose estimation task comprises two subtasks: \one segmenting the object from the scene, and \two predicting the orientation of the segmented object.
Therefore, the processing steps involves generating object masks for each label and augmenting the data to adapt to various environment in application.

\subsubsection{Generating masks}
\codename{} leverages the recent emergence of Segment-Anything Model (SAM) \cite{kirillov2023segment} which is capable of producing high quality segmentations given points or bounding boxes as prompts. 
For each image captured by \codename{}, the subsequent procedures must be executed:

\paragraph{Bounding box}
As the camera is calibrated to the robotic arm's coordinate frame, we generate the initial bounding box of the object by assuming the robotic arm is holding a 15x15x15 cm cube. We then project the cube's coordinates onto the camera's 2D plane to obtain the bounding box (Figure \ref{fig:processing}b). 
Generally, this method yields satisfactory masks for objects that are distinct and easily identifiable in the image. However, complications arise when objects are partially obscured by the robotic arm, difficult to distinguish (e.g., a flask whose appearance is influenced by the background), or even entirely invisible. To address these challenges, an additional filtering process is introduced. This process either segments the semi-occluded objects or discards the invisible data.

\paragraph{Filtering}
To improve the quality of the masked objects, we leverages the interaction with SAM by providing additional prompts (points) to specify the objects and background (Figure \ref{fig:processing}c). Specifically, we incorporated two additional steps: \one we provide additional prompts for the SAM to highlight the object's location and \two we wrap the gripper in green tape to reduce its potential interference with segmentation performance. 

\begin{itemize}
    \item \textbf{Providing additional prompts for the SAM to highlight the object's location}. Given that the gripper consistently holds the objects, we can infer that the center of the 15x15x15 cm cube corresponds to the object.Thus, we add the projected pixel coordinate of this center as a point prompt for the SAM, indicating the object's location.

    \item \textbf{Removing green background.} Given that the gripper can partially obscure the object, it might predominantly appear within the bounding box. This could lead the SAM to mistakenly segment the gripper as the target object. To counteract this, we detect the green regions in the image, which are presumed to represent the gripper. We then calculate the geometric center of these regions and use its coordinates to provide the SAM with a prompt, pointing out the undesired areas.
\end{itemize}

\subsubsection{Data augmentation}
% In our inference pipeline, we utilized a recent work named High-Quality Tracking Anything \cite{zhu2023tracking}. The result shows that it can produce high quality segmentation of the target object given a video input. 
We augment each masked image of the object with random exposure, contrast, saturation, etc. via Albumentations \cite{info11020125} to achieve better generalizability.

% Although data augmentation by pasting objects onto various backgrounds is feasible \cite{ghiasi2021simple, dwibedi2017cut}, we chose to adapt the data to environments in which potential users would need object pose estimation. This approach ensures that the augmented data is more representative of real-world scenarios and enhances the model's performance in relevant contexts.
% To do so, the user first captures a short video (\textasciitilde{}5 seconds) showing the target environment and hands movement. For each image of the data, \codename{} copy the segmented object (via the mask generated in the previous step) and paste to a random frame of the video.
% Additionally, we augment each image with random exposure, contrast, saturation, etc via Albumentations \cite{info11020125} achieve better generalizability.

\section{Evaluation}
% \xac{maybe rename it as `Results'?}

\label{sec:objects}

% In order to validate the proposed data collection pipeline, 
% we designed a simple pose estimation pipeline to testing on videos. The pipeline
% we conducted both quantitative and qualitative evaluation of the model trained on our collected data.
% Ideally, the model should be evaluated in the application setting, where the user will manipulating the object while estimating the pose. However,
% the challenge of annotating ground truth label still remains and annotating ground truth on video frames in which the object is being manipulated is extremely hard for human to manually label. 
% Therefore, 

% People may question the need for using a robotic arm to perform data collection, as there is recent work focusing on pose estimation of physical objects with few-shot learning approaches \cite{sun2022onepose, liu2022gen6d, wen2023bundlesdf}. 
% Typically, these methods require the input of a short video to reconstruct the 3D mesh of the objects \cite{sun2022onepose, liu2022gen6d} or a neural object field \cite{wen2023bundlesdf} to train on a small scale of data, and they have demonstrated competitive performance. 
% However, similar to other prior work, these methods do not address the pose estimation of appearance-changing objects, which is an essential aspect of our focus. 
% As a result, their performance may be limited when applied to objects with deformable properties, transparent materials, or multiple mechanical states. 
% Therefore, to justify the needs of \codename{} and showcase its potential, 
% 
%  To validate our proposed data collection pipeline, 
To demonstrate the feasibility of our data collection pipeline,
we conducted both quantitative and qualitative evaluation of the model trained on our data to compare with a few-shot learning pose estimation approach Gen6D \cite{liu2022gen6d}. Gen6D has shown competitive performance on any custom object by
using a single video as input for 3D reconstruction (via COLMAP~\cite{schonberger2016structure}) and performed feature matching based on the image and resulting pointcloud.

We evaluated the model in two settings, \textit{controlled} setting where the ground truth can be reliably obtained for quantitative evaluation, and \textit{application} setting, where the user manipulates the object during pose estimation for qualitative evaluation, as the ground truth pose cannot be obtained easily. 
% However, the challenge of ground truth annotation makes manual labeling of video frames, where the object is in motion and being manipulated, extremely difficult.
% Thus, we conducted a quantitative evaluation in the setting where the robotic arm actively manipulates and reposition the object. To ensure variation of the environment, we capture the testing data from different angles and backgrounds from the training phase, and the robotic arm operates in a distinct trajectory. 
% On the other hand, to test the model performance in the application setting, we conducted a qualitative evaluation in which the data consists of videos of user manipulating the objects
% On the other hand, to assess model performance in real-world scenarios, we conducted a qualitative evaluation using videos where users manipulate the objects. 

\begin{table*}[ht]
\centering

\renewcommand{\arraystretch}{1.1} % reduce the row height
\begin{tabular}{lcccccccc}
\hline

\hline

                      & anpanman & frog & pitcher & flask & bottle & scissors & clamp & spray \\

\hline

\textbf{\codename{}}  &    91.9  &  61.9    &   73.7    &  87.1(66.9) & 71.9  &  83.4 & 42.0  &  87.6 \\ 

Gen6D \cite{liu2022gen6d}  & 19.6 & 12.9 & 12.7 & 16.2 & 16.9 & 38.3 & 19.4 & 28.4 \\

\bottomrule

\multicolumn{9}{l}{\footnotesize *The number in parentheses indicates flask accuracy in a different background (Figure \ref{fig:quantitative}).}

\end{tabular}

\caption{Quantitative evaluation result. The numbers indicate the average precision at 20\degree~ azimuth error.}

\label{tab:quantitative}
% \xac{what do these numbers mean? explain in the caption}
\end{table*}

\subsection{Implementation of pose estimation pipeline}
Before we delve into the result of quantitative evaluation, We will discuss the pose estimation pipeline first.
As mentioned earlier, the pose estimation pipeline should consists of one model for segmenting the target object and another for predicting the orientation based on the segmented output. 
For objects with manually-modifiable states (e.g., scissors), an additional state classifier is employed.
% model to segment the target object and another model to predict the orientation of the object on the segmented object. 
% Additionally, if the object contains manually-changed states (\eg scissors), another state classifier is trained for that.
% For the segmentation, we utilized a recent work based on SAM named HQTrack \cite{zhu2023tracking}, which is able to produce high quality segmentation of the target object in any input video. 

For the segmentation task, we leveraged a recent advancement based on SAM: HQTrack \cite{zhu2023tracking}. It is a zero-shot approach and requires no training while being able to consistently produce high-quality segmentation of target objects in videos. 
% which is a zero-shot approach needless of training and is able to produce high-quality segmentation of target objects in any given video.

For the orientation estimation, the model is a VGG16 model, pretrained on ImageNet~\cite{imagenet}, followed by a fully connected layer outputting the quaternion and the 2D pixel location of the object.
The loss function is a combined loss of the Geodesic Loss on the quaternion prediction and the MSE Loss of the displacement prediction. 
We train the model on the augmented data for 120 epochs, using the Adam optimizer with a learning rate of 0.0001.

For the state classification, the model is a MobileNet V3~\cite{howard2019searching}, pretrained on ImageNet~\cite{imagenet}, followed by a fully connected layer, and the output dimension is equivalent to the number of the states of the object.
We train the model on the augmented data for 120 epochs, using the Adam optimizer with a learning rate of 0.0001.

\begin{figure}
    \centering
    \includegraphics[width=1\linewidth]{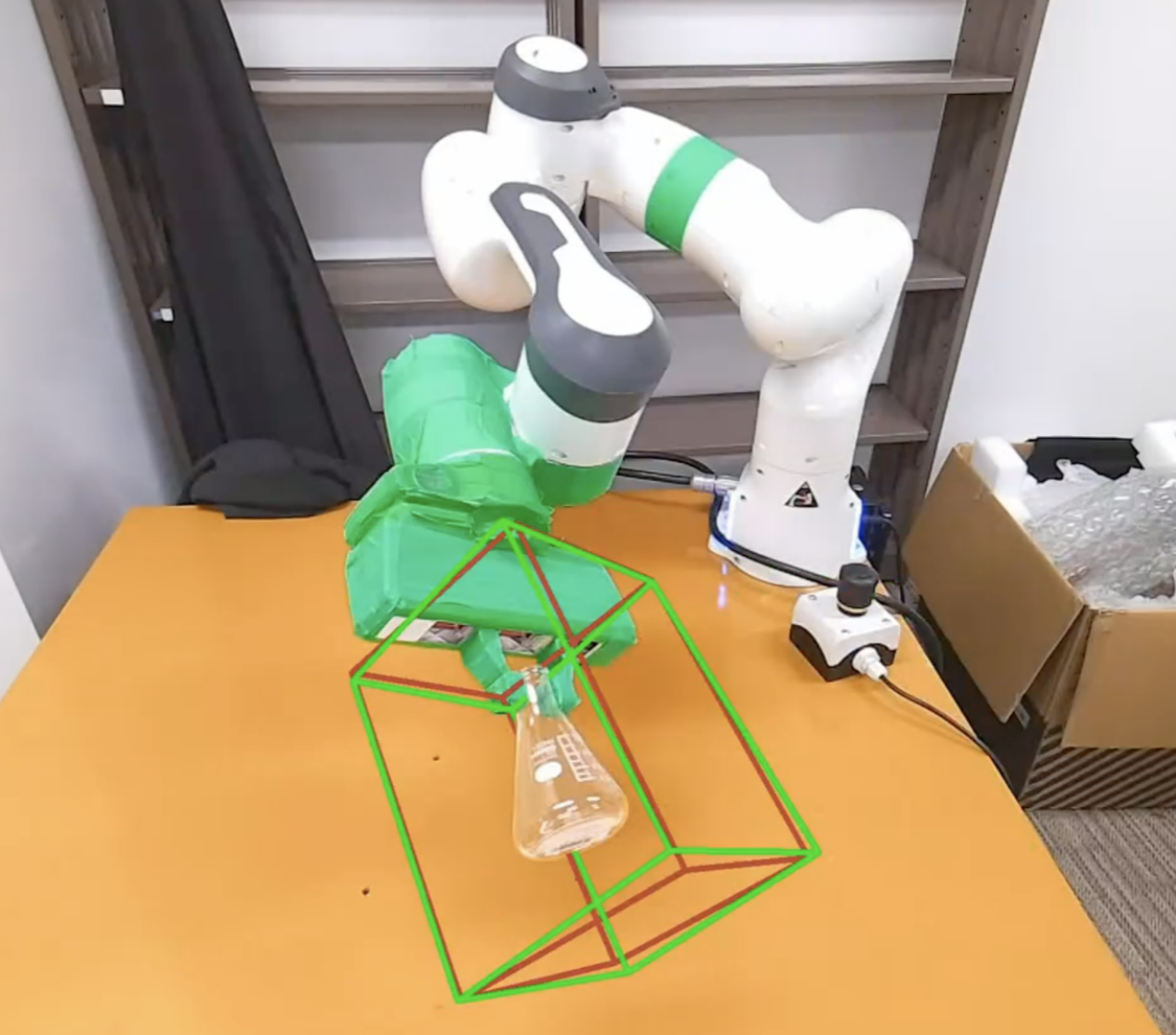}
    \caption{Quantitative evaluation setup. The green bounding box represents the ground truth and the read bounding box represents the predicted pose.
    }
    \label{fig:quantitative}
\end{figure}
\subsection{Quantitative evaluation}

For quantitative evaluation, we modified the environment by changing the camera angle and updating the background (Figure \ref{fig:quantitative}). Using a newly designed trajectory for the robotic arm, we sampled 1041 data entries. The accuracy threshold for pose estimation remained consistent with our training data parameters: set at 0.35 or an azimuth angle of 20\degree{}.

To clarify, our evaluation only focused on the accuracy of the orientation prediction, since \codename{} rely on prior to determine the position of the object.
For Gen6D, we could not modify its training pipeline to incorporate HQTrack to enhance its object detection. Instead, we adhered to the guidelines provided for pose estimation on custom objects as outlined in Gen6D's guidelines\footnote{\url{https://github.com/liuyuan-pal/Gen6D/blob/main/custom_object.md}}.
Specifically, we performed 3D reconstruction of the object using COLMAP \cite{schonberger2016structure} and followed the preprocessing procedure in the guideline.
% as we cannot decompose its training pipeline to insert HQTrack to enhance its object detection, we followed the guideline for performing pose estimation on custom object\footnote{https://github.com/liuyuan-pal/Gen6D/blob/main/custom_object.md}. 
% Specifically, we perform 3D reconstruction using COLMAP \cite{schonberger2016structure} and following the preprocessing procedure in the guideline.

% we executed the preprocessing procedure including 3D reconstruction of the object, and then evaluate the model using the test dataset. 

For objects with multiple manual states, test data is gathered for each state, and the model is evaluated accordingly. The results represent the mean accuracy across all states. Specifically, the flask was tested against two different backgrounds: its original setting (a black background) and an alternate setting with a typical orange-colored desk surface.
Table \ref{tab:quantitative} shows the accuracy comparison between our approach and Gen6D. 

We note that the accuracy is much lower in our testing result as compared to the result Gen6D demonstrated in their paper. 
This could be due to several factors:
\begin{itemize}
    \item 3D reconstruction failures (\eg Figure \ref{fig:3d}).
    \item Data collection with objects in static positions, leading to challenges when the object's unseen side becomes visible during manipulation (\eg a plush toy might be placed face-up on a table during data collection).
    \item Gen6D's documented issue with size-changing objects in frames (as the object moves closer and further away from the camera), as mentioned in their GitHub issues\footnote{https://github.com/liuyuan-pal/Gen6D/issues/29}.
    % There is a known issue of Gen6D reported by the authors in thier GitHub issues Gen6D may struggle to detect the object when the object changes size in the frames (as the object moves closer and further away from the camera).
\end{itemize}

The results indicate that a simple pose estimator trained with data from \codename{} can deliver relative working pose estimation performance. However, the quantitative findings also reveal some limitations.
For example, the accuracy of clamp is relatively low compared to other objects due to ambiguity caused by its symmetry. Additionally, objects whose appearances are environment-dependent demonstrate inconsistent performance under varying backgrounds. More details are discussed in the limitation in Sec. \ref{sec:limitations}.

\subsection{Qualitative evaluation}
\label{sec:qualitative}

\begin{figure*}
    \centering
    \includegraphics[width=\linewidth]{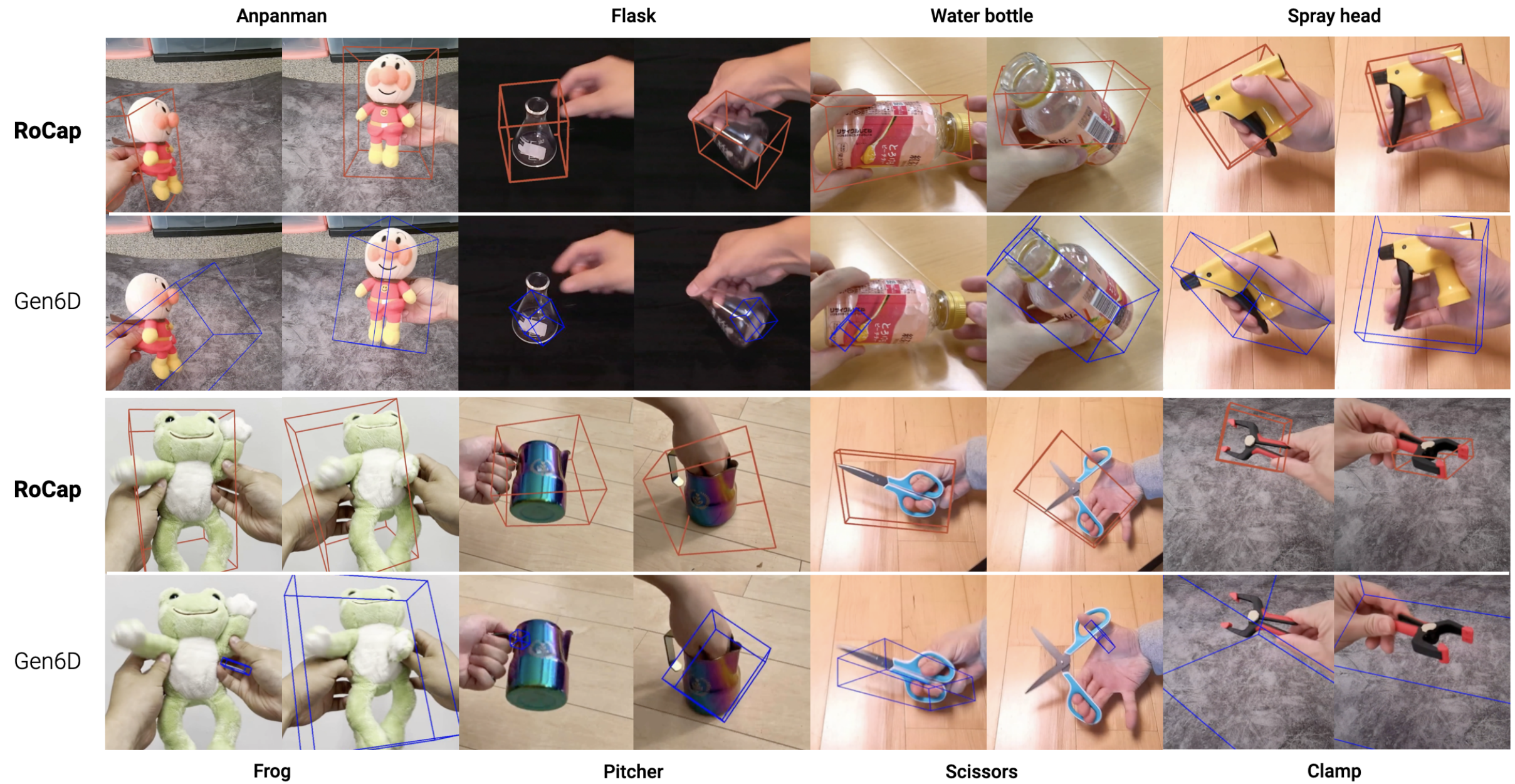}
    \caption{Qualitative evaluation of the eight objects.
    % Pose estimation of appearance-changing objects using model trained on data collected by \codename{} (top) and on 3D reconstructed data from Gen6D \cite{liu2022gen6d} (bottom). \nick{add another row of the new objects}
    }
    \label{fig:results}
\end{figure*}

To test the performance in the application setting, due to the difficulty in collecting ground truth, we conducted a qualitative evaluation on videos of humans manipulating the objects. Figure \ref{fig:results} shows the qualitative comparison between model trained on \codename{} data and Gen6D. 
For example, \codename{} recorded both closed and open states during data collection for the pair of scissors. This allowed it to provide viable pose estimation for the open state (Gen6D which struggled with the unobserved state).
Please refer to the supplementary materials for the video. 
\section{Discussion}

\subsection{Limitations}
\label{sec:limitations}

The quantitative and qualitative evaluation has demonstrated the feasibility and potential of our data collection method. However, the result also shows certain limitations. Below, we will discuss the limitations from the perspectives of data capturing, model performance and other constraints.

\begin{figure}
    \centering
    \includegraphics[width=1\linewidth]{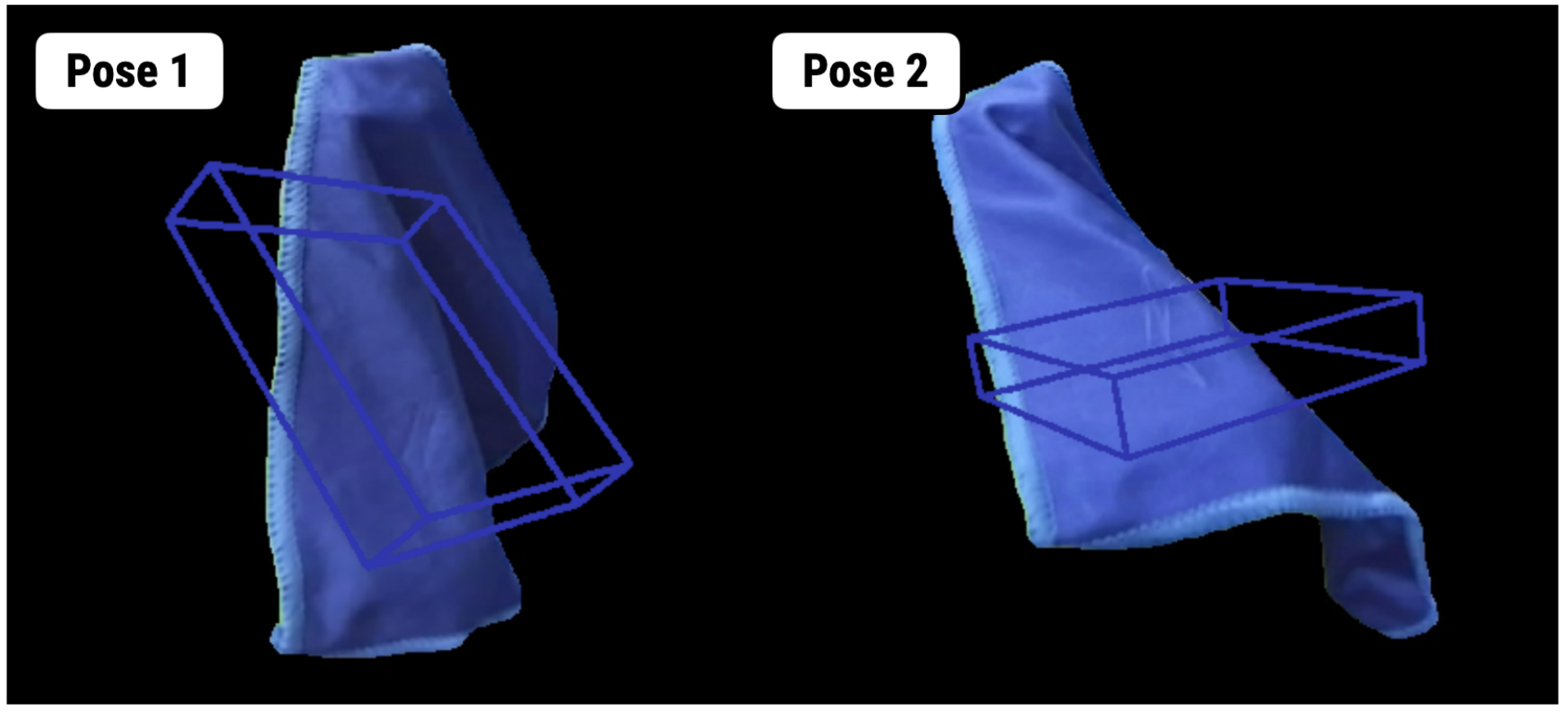}
    \caption{Failure case for a highly deformable cloth.
    }
    \label{fig:limitation}
\end{figure}

\paragraph{Data capturing}
While \codename{} addresses the data capturing of appearance-changing objects, it requires the objects have distinct appearances in different defined poses. One typically example that is challenging for \codename{} is cloth, which is highly deformable.
Its extreme flexibility results in a loss of the pose information when being manipulated by the robotic arm. As illustrated in Figure \ref{fig:limitation}, the piece of cloth is nearly identical in two different poses manipulated by the robotic arm.

On the other hand, as currently we target objects that people can easily change their appearances with hands, leading to the target object size ranging from 0.5x\textasciitilde1.5x of a palm size. Additionally, our robotic arm's mechanical gripper, with a maximum gripping width of 80mm, further constrains the size of objects it can handle.
However, this limitation can be resolved when a system applies our method to a larger scale robotic arm (\eg in a mass manufacturing setting).

\paragraph{Model performance}
Indicated in the evaluation results, the pose estimation pipeline does not handle symmetric object well. While this has been an open challenge in object pose estimation \cite{thalhammer2023challenges}, recent work have proposed different network architecture to address this issue \cite{xiang2017posecnn, song2020hybridpose}. While addressing symmetry is beyond the purview of this paper, future enhancements could incorporate a sophisticated pose estimator or gather supplementary data like depth via depth cameras.

Furthermore, variations in the environment from the capturing stage may also affect the model performance, especially for viewing-angle dependent objects. While it is feasible to maintain an environment similar to the capturing setup (\eg using a black background when operating a transparent flask), future improvements could include varying environmental factors such as different lighting conditions \cite{debevec2000acquiring, okabe2007light}. Additionally, future work could introduce other augmentation method such as \cite{zongker1999environment} to adapt to various environment. 

\paragraph{Other constraints}
Currently the need of a robotic arm may require a lab setting. However, the pipeline can also be applied to scenarios such as \one product manufacturers collect data and train a model for their product, and include it as part of their solution package and \two home users asks their robot to train a model for their own object when robots are more accessible in the future, and later the user uses the model to estimation the pose in specific applications.

% \zhongyi{}{Change the title into discussion}
% \subsection{Data Collection for More Appearance-Changing Features}
% At present, \codename{} takes into account appearance-changing features such as deformation, transparency, and multiple mechanical states. A future direction involves incorporating additional features to achieve a comprehensive coverage of appearance-changing objects. These features could originate from the objects themselves or from the environment.
% For instance, future data collection should consider object features like reflectivity (e.g., steel tweezers for medical surgery), texture variations, and objects with intricate structures. 
% Moreover, environmental factors such as varying lighting conditions \cite{debevec2000acquiring, okabe2007light}, different background should also be taken into account, as these factors can significantly influence the appearance of objects. Specifically, methods such as environment matting \cite{zongker1999environment} can be utilized to augment the data collection of transparent objects to adapt to a wider range of background.
% By expanding the scope of features considered in \codename{}, the collected data could potentially enhance the performance and robustness of existing deep learning models for pose estimation
% handling a wider range of real-world scenarios and complex objects.

\subsection{Handling Occlusion}
Occlusion happens in different scenarios, including the objects being manipulated by hands during interaction, or the objects being held by the robotic arm during data capturing. 
While the hand occlusion does have an impact on the model performance trained on \codename{} data, our pipeline is less impacted compared to Gen6D as shown in the qualitative results. This is due to the fact that during the capture phase, the robotic arm may partially obscure the object throughout the capturing process, which simulates hand occlusion in the training data. 

To further address the occlusion problem,
one possible approach is 
% to tackle occlusion caused by human hands during interaction is 
to introduce a hand-like robotic hand during the data capturing process. For example, anthropomorphic robotic hands, such as those presented in \cite{gama2014anthropomorphic}, can closely mimic human hand movements and provide more realistic interaction scenarios for data collection. By using a robotic hand, it is possible to better account for occlusions that occur during human-object interactions and develop models that can better predict user intent in such cases.
Additionally, to address occlusion caused by the robotic arm during data capturing, multiple cameras could be employed to ensure the complete visibility of the objects being captured.

\subsection{Automatic Changing of Mechanical States}
As mentioned in the paper, certain articulated objects necessitate human intervention to change their states, as they cannot be manipulated by the parallel gripper of the robotic arm \cite{li2022roman}. Examples of such objects include those that require a large range of motion or those that demand bi-manual operation.
% to change some of the mechanical states of the target appearance-changing objects, 
Recent research in HCI has proposed different methods of attaching mechanisms to the physical object to automatically actuate the motion without human intervention \cite{li2022roman, li2019robiot, li2020romeo}, which can be potentially leveraged by future data collection system using robotic arms to automatically collect a large amount of data.
By automating the data collection process, it is possible to scale up the dataset and sample object states at smaller intervals. For instance, instead of having discrete states of a clamp, we can sample from a continuous parameter space evenly while capturing. This would enable the prediction of the continuous parameter such as the angle of a pair of scissors, thus opening up a wider range of applications.
Future direction should include how to design mechanisms that will not affect the apperance of the objects during capturing while being able to actuate the objects.

% Incorporating these automated data collection methods in future iterations of \codename{} would not only increase the volume and variety of data available for training, but it would also enhance the system's ability to handle more complex scenarios and follow-up actions. As a result, the system's overall performance and applicability in real-world situations would be significantly improved.
% Using hand-shape end-effector or robotic hand during the capture

\subsection{Leveraging Robots for Large-Scale Data Collection}
Robots possess the capability to perform repetitive tasks consistently and efficiently. 
Researchers in computer vision and HCI have explored various approaches to employing robots for data collection across a diverse range of applications \cite{chong2021per, gao2022objectfolder, mandlekar2019scaling}. This has opened up new opportunities for augmenting tasks that necessitate a substantial amount of repetitive work, such as data collection for multiple objects, through the integration of robotic systems.
By leveraging robotic systems, researchers can not only streamline the data collection process but also minimize human error and fatigue. This can lead to the acquisition of more accurate and reliable datasets, which are critical for the development and evaluation of advanced algorithms and models. 

In addition to automating repetitive tasks, robotic systems can be equipped with various sensors and end effectors to collect multimodal data, such as visual, tactile, and auditory information. This can significantly enrich the datasets and provide researchers with a more comprehensive understanding of the objects and environments being studied.
As robotics technology continues to advance, we can expect even more sophisticated and versatile robotic systems to be employed in the data collection process. This will ultimately lead to more robust, accurate, and diverse datasets, which will contribute to the improvement of various computer vision and HCI applications.
\section{Conclusion}

In this paper, we present \codename{}, a robotic pipeline for data collection of appearance-changing objects. 
% This addresses the challenge of pose estimation for deformable objects (\eg plush toys), transparent objects (\eg glass flask), or objects with multiple mechanical states (\eg a clamp or a head of a spray bottle). 
% Using a robotic arm to hold such objects and capture the image data, we can train a simple deep learning model and perform pose estimation on the objects. 
% The results were compared with a state-of-the-art few-shot learning framework which was trained on 3D mesh reconstructed from a video.
% The comparison was rated by crowd workers and the results showed that the model trained on data collected by \codename{} outperforms the few-shot learning method, which shows the potential of \codename{}.
This system addresses the challenge of pose estimation for objects with deformable properties (\eg plush toys), viewing-angle dependent properties including transparent materials (\eg glass flasks) and reflective materials (\eg pitcher) or objects to be actuated with multiple mechanical states (\eg clamps or spray bottle heads). By employing a robotic arm to hold these objects and capture image data, which can be utilized by anyone possessing a 6 DoF robotic arm, we can train a simple deep learning model to perform pose estimation on the objects.
We conducted both quantitative and qualitative evaluation of our approach comparing to a few-shot learning framework, 
% compared the results of our approach with a state-of-the-art few-shot learning framework, 
which was trained on 3D mesh reconstructed from a video. 
The results demonstrated the feasibility and potential of \codename{}.
% The results demonstrated that the model trained on data collected by \codename{} significantly outperforms the few-shot learning method. This finding highlights the potential of \codename{} as an effective solution for pose estimation of appearance-changing objects, offering new possibilities in the realm of object recognition and manipulation tasks.

\begin{acks}
To Robert, for the bagels and explaining CMYK and color spaces.
\end{acks}

\bibliographystyle{ACM-Reference-Format}
\bibliography{reference}


\begin{thebibliography}{51}

%%% the Web address.

\ifx \showCODEN    \undefined \def \showCODEN     #1{\unskip}     \fi
\ifx \showDOI      \undefined \def \showDOI       #1{#1}\fi
\ifx \showISBNx    \undefined \def \showISBNx     #1{\unskip}     \fi
\ifx \showISBNxiii \undefined \def \showISBNxiii  #1{\unskip}     \fi
\ifx \showISSN     \undefined \def \showISSN      #1{\unskip}     \fi
\ifx \showLCCN     \undefined \def \showLCCN      #1{\unskip}     \fi
\ifx \shownote     \undefined \def \shownote      #1{#1}          \fi
\ifx \showarticletitle \undefined \def \showarticletitle #1{#1}   \fi
\ifx \showURL      \undefined \def \showURL       {\relax}        \fi
% invisible to TeX
\providecommand\bibfield[2]{#2}
\providecommand\bibinfo[2]{#2}
\providecommand\natexlab[1]{#1}
\providecommand\showeprint[2][]{arXiv:#2}

\bibitem[Ahmadyan et~al\mbox{.}(2021)]%
        {Ahmadyan2021objectron}
\bibfield{author}{\bibinfo{person}{Adel Ahmadyan}, \bibinfo{person}{Liangkai Zhang}, \bibinfo{person}{Artsiom Ablavatski}, \bibinfo{person}{Jianing Wei}, {and} \bibinfo{person}{Matthias Grundmann}.} \bibinfo{year}{2021}\natexlab{}.
\newblock \showarticletitle{Objectron: A Large Scale Dataset of Object-Centric Videos in the Wild With Pose Annotations}. In \bibinfo{booktitle}{\emph{Proceedings of the IEEE/CVF Conference on Computer Vision and Pattern Recognition (CVPR)}}. \bibinfo{pages}{7822--7831}.
\newblock


\bibitem[Barnes et~al\mbox{.}(2008)]%
        {barnes2008video}
\bibfield{author}{\bibinfo{person}{Connelly Barnes}, \bibinfo{person}{David~E Jacobs}, \bibinfo{person}{Jason Sanders}, \bibinfo{person}{Dan~B Goldman}, \bibinfo{person}{Szymon Rusinkiewicz}, \bibinfo{person}{Adam Finkelstein}, {and} \bibinfo{person}{Maneesh Agrawala}.} \bibinfo{year}{2008}\natexlab{}.
\newblock \showarticletitle{Video puppetry: a performative interface for cutout animation}.
\newblock In \bibinfo{booktitle}{\emph{ACM SIGGRAPH Asia 2008 papers}}. \bibinfo{pages}{1--9}.
\newblock


\bibitem[Brachmann et~al\mbox{.}(2014)]%
        {brachmann2014learning}
\bibfield{author}{\bibinfo{person}{Eric Brachmann}, \bibinfo{person}{Alexander Krull}, \bibinfo{person}{Frank Michel}, \bibinfo{person}{Stefan Gumhold}, \bibinfo{person}{Jamie Shotton}, {and} \bibinfo{person}{Carsten Rother}.} \bibinfo{year}{2014}\natexlab{}.
\newblock \showarticletitle{Learning 6d object pose estimation using 3d object coordinates}. In \bibinfo{booktitle}{\emph{European conference on computer vision}}. Springer, \bibinfo{pages}{536--551}.
\newblock


\bibitem[Buslaev et~al\mbox{.}(2020)]%
        {info11020125}
\bibfield{author}{\bibinfo{person}{Alexander Buslaev}, \bibinfo{person}{Vladimir~I. Iglovikov}, \bibinfo{person}{Eugene Khvedchenya}, \bibinfo{person}{Alex Parinov}, \bibinfo{person}{Mikhail Druzhinin}, {and} \bibinfo{person}{Alexandr~A. Kalinin}.} \bibinfo{year}{2020}\natexlab{}.
\newblock \showarticletitle{Albumentations: Fast and Flexible Image Augmentations}.
\newblock \bibinfo{journal}{\emph{Information}} \bibinfo{volume}{11}, \bibinfo{number}{2} (\bibinfo{year}{2020}).
\newblock
\showISSN{2078-2489}
\urldef\tempurl%
\url{https://doi.org/10.3390/info11020125}
\showDOI{\tempurl}


\bibitem[Calli et~al\mbox{.}(2015)]%
        {calli2015ycb}
\bibfield{author}{\bibinfo{person}{Berk Calli}, \bibinfo{person}{Arjun Singh}, \bibinfo{person}{Aaron Walsman}, \bibinfo{person}{Siddhartha Srinivasa}, \bibinfo{person}{Pieter Abbeel}, {and} \bibinfo{person}{Aaron~M Dollar}.} \bibinfo{year}{2015}\natexlab{}.
\newblock \showarticletitle{The ycb object and model set: Towards common benchmarks for manipulation research}. In \bibinfo{booktitle}{\emph{2015 international conference on advanced robotics (ICAR)}}. IEEE, \bibinfo{pages}{510--517}.
\newblock


\bibitem[Chao et~al\mbox{.}(2021)]%
        {chao2021dexycb}
\bibfield{author}{\bibinfo{person}{Yu-Wei Chao}, \bibinfo{person}{Wei Yang}, \bibinfo{person}{Yu Xiang}, \bibinfo{person}{Pavlo Molchanov}, \bibinfo{person}{Ankur Handa}, \bibinfo{person}{Jonathan Tremblay}, \bibinfo{person}{Yashraj~S Narang}, \bibinfo{person}{Karl Van~Wyk}, \bibinfo{person}{Umar Iqbal}, \bibinfo{person}{Stan Birchfield}, {et~al\mbox{.}}} \bibinfo{year}{2021}\natexlab{}.
\newblock \showarticletitle{DexYCB: A benchmark for capturing hand grasping of objects}. In \bibinfo{booktitle}{\emph{Proceedings of the IEEE/CVF Conference on Computer Vision and Pattern Recognition}}. \bibinfo{pages}{9044--9053}.
\newblock


\bibitem[Chong et~al\mbox{.}(2021)]%
        {chong2021per}
\bibfield{author}{\bibinfo{person}{Toby Chong}, \bibinfo{person}{I-Chao Shen}, \bibinfo{person}{Nobuyuki Umetani}, {and} \bibinfo{person}{Takeo Igarashi}.} \bibinfo{year}{2021}\natexlab{}.
\newblock \showarticletitle{Per Garment Capture and Synthesis for Real-time Virtual Try-on}. In \bibinfo{booktitle}{\emph{The 34th Annual ACM Symposium on User Interface Software and Technology}}. \bibinfo{pages}{457--469}.
\newblock


\bibitem[Danila~Rukhovich(2022)]%
        {rukhovich2021imvoxelnet}
\bibfield{author}{\bibinfo{person}{Anton~Konushin Danila~Rukhovich, Anna~Vorontsova}.} \bibinfo{year}{2022}\natexlab{}.
\newblock \showarticletitle{ImVoxelNet: Image to Voxels Projection for Monocular and Multi-View General-Purpose 3D Object Detection}.
\newblock  (\bibinfo{year}{2022}), \bibinfo{pages}{2397--2406}.
\newblock


\bibitem[Debevec et~al\mbox{.}(2000)]%
        {debevec2000acquiring}
\bibfield{author}{\bibinfo{person}{Paul Debevec}, \bibinfo{person}{Tim Hawkins}, \bibinfo{person}{Chris Tchou}, \bibinfo{person}{Haarm-Pieter Duiker}, \bibinfo{person}{Westley Sarokin}, {and} \bibinfo{person}{Mark Sagar}.} \bibinfo{year}{2000}\natexlab{}.
\newblock \showarticletitle{Acquiring the reflectance field of a human face}. In \bibinfo{booktitle}{\emph{Proceedings of the 27th annual conference on Computer graphics and interactive techniques}}. \bibinfo{pages}{145--156}.
\newblock


\bibitem[Denavit and Hartenberg(1955)]%
        {denavit1955kinematic}
\bibfield{author}{\bibinfo{person}{Jacques Denavit} {and} \bibinfo{person}{Richard~S Hartenberg}.} \bibinfo{year}{1955}\natexlab{}.
\newblock \showarticletitle{A kinematic notation for lower-pair mechanisms based on matrices}.
\newblock  (\bibinfo{year}{1955}).
\newblock


\bibitem[Deng et~al\mbox{.}(2009)]%
        {imagenet}
\bibfield{author}{\bibinfo{person}{Jia Deng}, \bibinfo{person}{Wei Dong}, \bibinfo{person}{Richard Socher}, \bibinfo{person}{Li-Jia Li}, \bibinfo{person}{Kai Li}, {and} \bibinfo{person}{Li Fei-Fei}.} \bibinfo{year}{2009}\natexlab{}.
\newblock \showarticletitle{ImageNet: A large-scale hierarchical image database}. In \bibinfo{booktitle}{\emph{2009 IEEE Conference on Computer Vision and Pattern Recognition}}. \bibinfo{pages}{248--255}.
\newblock
\urldef\tempurl%
\url{https://doi.org/10.1109/CVPR.2009.5206848}
\showDOI{\tempurl}


\bibitem[Dogan et~al\mbox{.}(2022)]%
        {dogan2022infraredtags}
\bibfield{author}{\bibinfo{person}{Mustafa~Doga Dogan}, \bibinfo{person}{Ahmad Taka}, \bibinfo{person}{Michael Lu}, \bibinfo{person}{Yunyi Zhu}, \bibinfo{person}{Akshat Kumar}, \bibinfo{person}{Aakar Gupta}, {and} \bibinfo{person}{Stefanie Mueller}.} \bibinfo{year}{2022}\natexlab{}.
\newblock \showarticletitle{InfraredTags: Embedding Invisible AR Markers and Barcodes Using Low-Cost, Infrared-Based 3D Printing and Imaging Tools}. In \bibinfo{booktitle}{\emph{CHI Conference on Human Factors in Computing Systems}}. \bibinfo{pages}{1--12}.
\newblock


\bibitem[Dwibedi et~al\mbox{.}(2017)]%
        {dwibedi2017cut}
\bibfield{author}{\bibinfo{person}{Debidatta Dwibedi}, \bibinfo{person}{Ishan Misra}, {and} \bibinfo{person}{Martial Hebert}.} \bibinfo{year}{2017}\natexlab{}.
\newblock \showarticletitle{Cut, paste and learn: Surprisingly easy synthesis for instance detection}. In \bibinfo{booktitle}{\emph{Proceedings of the IEEE international conference on computer vision}}. \bibinfo{pages}{1301--1310}.
\newblock


\bibitem[Fan et~al\mbox{.}(2021)]%
        {fan2021transparent}
\bibfield{author}{\bibinfo{person}{Heng Fan}, \bibinfo{person}{Halady~Akhilesha Miththanthaya}, \bibinfo{person}{Siranjiv~Ramana Rajan}, \bibinfo{person}{Xiaoqiong Liu}, \bibinfo{person}{Zhilin Zou}, \bibinfo{person}{Yuewei Lin}, \bibinfo{person}{Haibin Ling}, {et~al\mbox{.}}} \bibinfo{year}{2021}\natexlab{}.
\newblock \showarticletitle{Transparent object tracking benchmark}. In \bibinfo{booktitle}{\emph{Proceedings of the IEEE/CVF International Conference on Computer Vision}}. \bibinfo{pages}{10734--10743}.
\newblock


\bibitem[Gama~Melo et~al\mbox{.}(2014)]%
        {gama2014anthropomorphic}
\bibfield{author}{\bibinfo{person}{Erika~Nathalia Gama~Melo}, \bibinfo{person}{Oscar~Fernando Aviles~Sanchez}, {and} \bibinfo{person}{Darlo Amaya~Hurtado}.} \bibinfo{year}{2014}\natexlab{}.
\newblock \showarticletitle{Anthropomorphic robotic hands: a review}.
\newblock \bibinfo{journal}{\emph{Ingenier{\'\i}a y desarrollo}} \bibinfo{volume}{32}, \bibinfo{number}{2} (\bibinfo{year}{2014}), \bibinfo{pages}{279--313}.
\newblock


\bibitem[Gao et~al\mbox{.}(2022)]%
        {gao2022objectfolder}
\bibfield{author}{\bibinfo{person}{Ruohan Gao}, \bibinfo{person}{Zilin Si}, \bibinfo{person}{Yen-Yu Chang}, \bibinfo{person}{Samuel Clarke}, \bibinfo{person}{Jeannette Bohg}, \bibinfo{person}{Li Fei-Fei}, \bibinfo{person}{Wenzhen Yuan}, {and} \bibinfo{person}{Jiajun Wu}.} \bibinfo{year}{2022}\natexlab{}.
\newblock \showarticletitle{Objectfolder 2.0: A multisensory object dataset for sim2real transfer}. In \bibinfo{booktitle}{\emph{Proceedings of the IEEE/CVF conference on computer vision and pattern recognition}}. \bibinfo{pages}{10598--10608}.
\newblock


\bibitem[Garrido-Jurado et~al\mbox{.}(2014)]%
        {garrido2014automatic}
\bibfield{author}{\bibinfo{person}{Sergio Garrido-Jurado}, \bibinfo{person}{Rafael Mu{\~n}oz-Salinas}, \bibinfo{person}{Francisco~Jos{\'e} Madrid-Cuevas}, {and} \bibinfo{person}{Manuel~Jes{\'u}s Mar{\'\i}n-Jim{\'e}nez}.} \bibinfo{year}{2014}\natexlab{}.
\newblock \showarticletitle{Automatic generation and detection of highly reliable fiducial markers under occlusion}.
\newblock \bibinfo{journal}{\emph{Pattern Recognition}} \bibinfo{volume}{47}, \bibinfo{number}{6} (\bibinfo{year}{2014}), \bibinfo{pages}{2280--2292}.
\newblock


\bibitem[Hampali et~al\mbox{.}(2020)]%
        {hampali2020honnotate}
\bibfield{author}{\bibinfo{person}{Shreyas Hampali}, \bibinfo{person}{Mahdi Rad}, \bibinfo{person}{Markus Oberweger}, {and} \bibinfo{person}{Vincent Lepetit}.} \bibinfo{year}{2020}\natexlab{}.
\newblock \showarticletitle{Honnotate: A method for 3d annotation of hand and object poses}. In \bibinfo{booktitle}{\emph{Proceedings of the IEEE/CVF conference on computer vision and pattern recognition}}. \bibinfo{pages}{3196--3206}.
\newblock


\bibitem[Held et~al\mbox{.}(2012)]%
        {held2012puppetry}
\bibfield{author}{\bibinfo{person}{Robert Held}, \bibinfo{person}{Ankit Gupta}, \bibinfo{person}{Brian Curless}, {and} \bibinfo{person}{Maneesh Agrawala}.} \bibinfo{year}{2012}\natexlab{}.
\newblock \showarticletitle{3D Puppetry: A Kinect-Based Interface for 3D Animation}. In \bibinfo{booktitle}{\emph{Proceedings of the 25th Annual ACM Symposium on User Interface Software and Technology}} (Cambridge, Massachusetts, USA) \emph{(\bibinfo{series}{UIST '12})}. \bibinfo{publisher}{Association for Computing Machinery}, \bibinfo{address}{New York, NY, USA}, \bibinfo{pages}{423–434}.
\newblock
\showISBNx{9781450315807}
\urldef\tempurl%
\url{https://doi.org/10.1145/2380116.2380170}
\showDOI{\tempurl}


\bibitem[Hettiarachchi and Wigdor(2016)]%
        {hettiarachchi2016annexing}
\bibfield{author}{\bibinfo{person}{Anuruddha Hettiarachchi} {and} \bibinfo{person}{Daniel Wigdor}.} \bibinfo{year}{2016}\natexlab{}.
\newblock \showarticletitle{Annexing reality: Enabling opportunistic use of everyday objects as tangible proxies in augmented reality}. In \bibinfo{booktitle}{\emph{Proceedings of the 2016 CHI Conference on Human Factors in Computing Systems}}. \bibinfo{pages}{1957--1967}.
\newblock


\bibitem[Hinterstoisser et~al\mbox{.}(2013)]%
        {hinterstoisser2013model}
\bibfield{author}{\bibinfo{person}{Stefan Hinterstoisser}, \bibinfo{person}{Vincent Lepetit}, \bibinfo{person}{Slobodan Ilic}, \bibinfo{person}{Stefan Holzer}, \bibinfo{person}{Gary Bradski}, \bibinfo{person}{Kurt Konolige}, {and} \bibinfo{person}{Nassir Navab}.} \bibinfo{year}{2013}\natexlab{}.
\newblock \showarticletitle{Model based training, detection and pose estimation of texture-less 3d objects in heavily cluttered scenes}. In \bibinfo{booktitle}{\emph{Computer Vision--ACCV 2012: 11th Asian Conference on Computer Vision, Daejeon, Korea, November 5-9, 2012, Revised Selected Papers, Part I 11}}. Springer, \bibinfo{pages}{548--562}.
\newblock


\bibitem[Hodan et~al\mbox{.}(2017)]%
        {hodan2017t}
\bibfield{author}{\bibinfo{person}{Tom{\'a}{\v{s}} Hodan}, \bibinfo{person}{Pavel Haluza}, \bibinfo{person}{{\v{S}}tep{\'a}n Obdr{\v{z}}{\'a}lek}, \bibinfo{person}{Jiri Matas}, \bibinfo{person}{Manolis Lourakis}, {and} \bibinfo{person}{Xenophon Zabulis}.} \bibinfo{year}{2017}\natexlab{}.
\newblock \showarticletitle{T-LESS: An RGB-D dataset for 6D pose estimation of texture-less objects}. In \bibinfo{booktitle}{\emph{2017 IEEE Winter Conference on Applications of Computer Vision (WACV)}}. IEEE, \bibinfo{pages}{880--888}.
\newblock


\bibitem[Howard et~al\mbox{.}(2019)]%
        {howard2019searching}
\bibfield{author}{\bibinfo{person}{Andrew Howard}, \bibinfo{person}{Mark Sandler}, \bibinfo{person}{Grace Chu}, \bibinfo{person}{Liang-Chieh Chen}, \bibinfo{person}{Bo Chen}, \bibinfo{person}{Mingxing Tan}, \bibinfo{person}{Weijun Wang}, \bibinfo{person}{Yukun Zhu}, \bibinfo{person}{Ruoming Pang}, \bibinfo{person}{Vijay Vasudevan}, \bibinfo{person}{Quoc~V. Le}, {and} \bibinfo{person}{Hartwig Adam}.} \bibinfo{year}{2019}\natexlab{}.
\newblock \bibinfo{title}{Searching for MobileNetV3}.
\newblock
\newblock
\showeprint[arxiv]{1905.02244}~[cs.CV]


\bibitem[Kalaitzakis et~al\mbox{.}(2021)]%
        {kalaitzakis2021fiducial}
\bibfield{author}{\bibinfo{person}{Michail Kalaitzakis}, \bibinfo{person}{Brennan Cain}, \bibinfo{person}{Sabrina Carroll}, \bibinfo{person}{Anand Ambrosi}, \bibinfo{person}{Camden Whitehead}, {and} \bibinfo{person}{Nikolaos Vitzilaios}.} \bibinfo{year}{2021}\natexlab{}.
\newblock \showarticletitle{Fiducial markers for pose estimation}.
\newblock \bibinfo{journal}{\emph{Journal of Intelligent \& Robotic Systems}} \bibinfo{volume}{101}, \bibinfo{number}{4} (\bibinfo{year}{2021}), \bibinfo{pages}{1--26}.
\newblock


\bibitem[Kirillov et~al\mbox{.}(2023)]%
        {kirillov2023segment}
\bibfield{author}{\bibinfo{person}{Alexander Kirillov}, \bibinfo{person}{Eric Mintun}, \bibinfo{person}{Nikhila Ravi}, \bibinfo{person}{Hanzi Mao}, \bibinfo{person}{Chloe Rolland}, \bibinfo{person}{Laura Gustafson}, \bibinfo{person}{Tete Xiao}, \bibinfo{person}{Spencer Whitehead}, \bibinfo{person}{Alexander~C Berg}, \bibinfo{person}{Wan-Yen Lo}, {et~al\mbox{.}}} \bibinfo{year}{2023}\natexlab{}.
\newblock \showarticletitle{Segment anything}.
\newblock \bibinfo{journal}{\emph{arXiv preprint arXiv:2304.02643}} (\bibinfo{year}{2023}).
\newblock


\bibitem[Krull et~al\mbox{.}(2015)]%
        {krull2015learning}
\bibfield{author}{\bibinfo{person}{Alexander Krull}, \bibinfo{person}{Eric Brachmann}, \bibinfo{person}{Frank Michel}, \bibinfo{person}{Michael~Ying Yang}, \bibinfo{person}{Stefan Gumhold}, {and} \bibinfo{person}{Carsten Rother}.} \bibinfo{year}{2015}\natexlab{}.
\newblock \showarticletitle{Learning analysis-by-synthesis for 6D pose estimation in RGB-D images}. In \bibinfo{booktitle}{\emph{Proceedings of the IEEE international conference on computer vision}}. \bibinfo{pages}{954--962}.
\newblock


\bibitem[Li et~al\mbox{.}(2020a)]%
        {li2020romeo}
\bibfield{author}{\bibinfo{person}{Jiahao Li}, \bibinfo{person}{Meilin Cui}, \bibinfo{person}{Jeeeun Kim}, {and} \bibinfo{person}{Xiang'Anthony' Chen}.} \bibinfo{year}{2020}\natexlab{a}.
\newblock \showarticletitle{Romeo: A design tool for embedding transformable parts in 3d models to robotically augment default functionalities}. In \bibinfo{booktitle}{\emph{Proceedings of the 33rd Annual Acm Symposium on User Interface Software and Technology}}. \bibinfo{pages}{897--911}.
\newblock


\bibitem[Li et~al\mbox{.}(2019)]%
        {li2019robiot}
\bibfield{author}{\bibinfo{person}{Jiahao Li}, \bibinfo{person}{Jeeeun Kim}, {and} \bibinfo{person}{Xiang'Anthony' Chen}.} \bibinfo{year}{2019}\natexlab{}.
\newblock \showarticletitle{Robiot: A design tool for actuating everyday objects with automatically generated 3D printable mechanisms}. In \bibinfo{booktitle}{\emph{Proceedings of the 32nd Annual ACM Symposium on User Interface Software and Technology}}. \bibinfo{pages}{673--685}.
\newblock


\bibitem[Li et~al\mbox{.}(2022)]%
        {li2022roman}
\bibfield{author}{\bibinfo{person}{Jiahao Li}, \bibinfo{person}{Alexis Samoylov}, \bibinfo{person}{Jeeeun Kim}, {and} \bibinfo{person}{Xiang~'Anthony' Chen}.} \bibinfo{year}{2022}\natexlab{}.
\newblock \showarticletitle{Roman: Making Everyday Objects Robotically Manipulable with 3D-Printable Add-on Mechanisms}. In \bibinfo{booktitle}{\emph{Proceedings of the 2022 CHI Conference on Human Factors in Computing Systems}} (New Orleans, LA, USA) \emph{(\bibinfo{series}{CHI '22})}. \bibinfo{publisher}{Association for Computing Machinery}, \bibinfo{address}{New York, NY, USA}, Article \bibinfo{articleno}{272}, \bibinfo{numpages}{17}~pages.
\newblock
\showISBNx{9781450391573}
\urldef\tempurl%
\url{https://doi.org/10.1145/3491102.3501818}
\showDOI{\tempurl}


\bibitem[Li et~al\mbox{.}(2020b)]%
        {li2020category}
\bibfield{author}{\bibinfo{person}{Xiaolong Li}, \bibinfo{person}{He Wang}, \bibinfo{person}{Li Yi}, \bibinfo{person}{Leonidas~J Guibas}, \bibinfo{person}{A~Lynn Abbott}, {and} \bibinfo{person}{Shuran Song}.} \bibinfo{year}{2020}\natexlab{b}.
\newblock \showarticletitle{Category-level articulated object pose estimation}. In \bibinfo{booktitle}{\emph{Proceedings of the IEEE/CVF conference on computer vision and pattern recognition}}. \bibinfo{pages}{3706--3715}.
\newblock


\bibitem[Liu et~al\mbox{.}(2021)]%
        {liu2021semi}
\bibfield{author}{\bibinfo{person}{Shaowei Liu}, \bibinfo{person}{Hanwen Jiang}, \bibinfo{person}{Jiarui Xu}, \bibinfo{person}{Sifei Liu}, {and} \bibinfo{person}{Xiaolong Wang}.} \bibinfo{year}{2021}\natexlab{}.
\newblock \showarticletitle{Semi-supervised 3d hand-object poses estimation with interactions in time}. In \bibinfo{booktitle}{\emph{Proceedings of the IEEE/CVF Conference on Computer Vision and Pattern Recognition}}. \bibinfo{pages}{14687--14697}.
\newblock


\bibitem[Liu et~al\mbox{.}(2022)]%
        {liu2022gen6d}
\bibfield{author}{\bibinfo{person}{Yuan Liu}, \bibinfo{person}{Yilin Wen}, \bibinfo{person}{Sida Peng}, \bibinfo{person}{Cheng Lin}, \bibinfo{person}{Xiaoxiao Long}, \bibinfo{person}{Taku Komura}, {and} \bibinfo{person}{Wenping Wang}.} \bibinfo{year}{2022}\natexlab{}.
\newblock \showarticletitle{Gen6D: Generalizable Model-Free 6-DoF Object Pose Estimation from RGB Images}.
\newblock \bibinfo{journal}{\emph{arXiv preprint arXiv:2204.10776}} (\bibinfo{year}{2022}).
\newblock


\bibitem[Lugaresi et~al\mbox{.}(2019)]%
        {lugaresi2019mediapipe}
\bibfield{author}{\bibinfo{person}{Camillo Lugaresi}, \bibinfo{person}{Jiuqiang Tang}, \bibinfo{person}{Hadon Nash}, \bibinfo{person}{Chris McClanahan}, \bibinfo{person}{Esha Uboweja}, \bibinfo{person}{Michael Hays}, \bibinfo{person}{Fan Zhang}, \bibinfo{person}{Chuo-Ling Chang}, \bibinfo{person}{Ming~Guang Yong}, \bibinfo{person}{Juhyun Lee}, \bibinfo{person}{Wan-Teh Chang}, \bibinfo{person}{Wei Hua}, \bibinfo{person}{Manfred Georg}, {and} \bibinfo{person}{Matthias Grundmann}.} \bibinfo{year}{2019}\natexlab{}.
\newblock \bibinfo{title}{MediaPipe: A Framework for Building Perception Pipelines}.
\newblock
\newblock
\urldef\tempurl%
\url{https://doi.org/10.48550/ARXIV.1906.08172}
\showDOI{\tempurl}


\bibitem[Mandlekar et~al\mbox{.}(2019)]%
        {mandlekar2019scaling}
\bibfield{author}{\bibinfo{person}{Ajay Mandlekar}, \bibinfo{person}{Jonathan Booher}, \bibinfo{person}{Max Spero}, \bibinfo{person}{Albert Tung}, \bibinfo{person}{Anchit Gupta}, \bibinfo{person}{Yuke Zhu}, \bibinfo{person}{Animesh Garg}, \bibinfo{person}{Silvio Savarese}, {and} \bibinfo{person}{Li Fei-Fei}.} \bibinfo{year}{2019}\natexlab{}.
\newblock \showarticletitle{Scaling robot supervision to hundreds of hours with roboturk: Robotic manipulation dataset through human reasoning and dexterity}. In \bibinfo{booktitle}{\emph{2019 IEEE/RSJ International Conference on Intelligent Robots and Systems (IROS)}}. IEEE, \bibinfo{pages}{1048--1055}.
\newblock


\bibitem[Marion et~al\mbox{.}(2018)]%
        {marion2018label}
\bibfield{author}{\bibinfo{person}{Pat Marion}, \bibinfo{person}{Peter~R Florence}, \bibinfo{person}{Lucas Manuelli}, {and} \bibinfo{person}{Russ Tedrake}.} \bibinfo{year}{2018}\natexlab{}.
\newblock \showarticletitle{Label fusion: A pipeline for generating ground truth labels for real rgbd data of cluttered scenes}. In \bibinfo{booktitle}{\emph{2018 IEEE International Conference on Robotics and Automation (ICRA)}}. IEEE, \bibinfo{pages}{3235--3242}.
\newblock


\bibitem[Okabe et~al\mbox{.}(2007)]%
        {okabe2007light}
\bibfield{author}{\bibinfo{person}{Makoto Okabe}, \bibinfo{person}{Kenshi Takayama}, \bibinfo{person}{Takashi Ijiri}, {and} \bibinfo{person}{Takeo Igarashi}.} \bibinfo{year}{2007}\natexlab{}.
\newblock \showarticletitle{Light shower: a poor man's light stage built with an off-the-shelf umbrella and projector}.
\newblock In \bibinfo{booktitle}{\emph{ACM SIGGRAPH 2007 sketches}}. \bibinfo{pages}{62--es}.
\newblock


\bibitem[Park and Martin(1994)]%
        {park1994robot}
\bibfield{author}{\bibinfo{person}{Frank~C Park} {and} \bibinfo{person}{Bryan~J Martin}.} \bibinfo{year}{1994}\natexlab{}.
\newblock \showarticletitle{Robot sensor calibration: solving AX= XB on the Euclidean group}.
\newblock \bibinfo{journal}{\emph{IEEE Transactions on Robotics and Automation}} \bibinfo{volume}{10}, \bibinfo{number}{5} (\bibinfo{year}{1994}), \bibinfo{pages}{717--721}.
\newblock


\bibitem[Peng et~al\mbox{.}(2019)]%
        {peng2019pvnet}
\bibfield{author}{\bibinfo{person}{Sida Peng}, \bibinfo{person}{Yuan Liu}, \bibinfo{person}{Qixing Huang}, \bibinfo{person}{Xiaowei Zhou}, {and} \bibinfo{person}{Hujun Bao}.} \bibinfo{year}{2019}\natexlab{}.
\newblock \showarticletitle{Pvnet: Pixel-wise voting network for 6dof pose estimation}. In \bibinfo{booktitle}{\emph{Proceedings of the IEEE/CVF Conference on Computer Vision and Pattern Recognition}}. \bibinfo{pages}{4561--4570}.
\newblock


\bibitem[Qian et~al\mbox{.}(2022)]%
        {qian2022arnnotate}
\bibfield{author}{\bibinfo{person}{Xun Qian}, \bibinfo{person}{Fengming He}, \bibinfo{person}{Xiyun Hu}, \bibinfo{person}{Tianyi Wang}, {and} \bibinfo{person}{Karthik Ramani}.} \bibinfo{year}{2022}\natexlab{}.
\newblock \showarticletitle{ARnnotate: An Augmented Reality Interface for Collecting Custom Dataset of 3D Hand-Object Interaction Pose Estimation}. In \bibinfo{booktitle}{\emph{Proceedings of the 35th Annual ACM Symposium on User Interface Software and Technology}}. \bibinfo{pages}{1--14}.
\newblock


\bibitem[Ros et~al\mbox{.}(2016)]%
        {ros2016synthia}
\bibfield{author}{\bibinfo{person}{German Ros}, \bibinfo{person}{Laura Sellart}, \bibinfo{person}{Joanna Materzynska}, \bibinfo{person}{David Vazquez}, {and} \bibinfo{person}{Antonio~M Lopez}.} \bibinfo{year}{2016}\natexlab{}.
\newblock \showarticletitle{The synthia dataset: A large collection of synthetic images for semantic segmentation of urban scenes}. In \bibinfo{booktitle}{\emph{Proceedings of the IEEE conference on computer vision and pattern recognition}}. \bibinfo{pages}{3234--3243}.
\newblock


\bibitem[Schonberger and Frahm(2016)]%
        {schonberger2016structure}
\bibfield{author}{\bibinfo{person}{Johannes~L Schonberger} {and} \bibinfo{person}{Jan-Michael Frahm}.} \bibinfo{year}{2016}\natexlab{}.
\newblock \showarticletitle{Structure-from-motion revisited}. In \bibinfo{booktitle}{\emph{Proceedings of the IEEE conference on computer vision and pattern recognition}}. \bibinfo{pages}{4104--4113}.
\newblock


\bibitem[Song et~al\mbox{.}(2020)]%
        {song2020hybridpose}
\bibfield{author}{\bibinfo{person}{Chen Song}, \bibinfo{person}{Jiaru Song}, {and} \bibinfo{person}{Qixing Huang}.} \bibinfo{year}{2020}\natexlab{}.
\newblock \showarticletitle{Hybridpose: 6d object pose estimation under hybrid representations}. In \bibinfo{booktitle}{\emph{Proceedings of the IEEE/CVF conference on computer vision and pattern recognition}}. \bibinfo{pages}{431--440}.
\newblock


\bibitem[Su et~al\mbox{.}(2021)]%
        {su2021synpo}
\bibfield{author}{\bibinfo{person}{Yongzhi Su}, \bibinfo{person}{Jason Rambach}, \bibinfo{person}{Alain Pagani}, {and} \bibinfo{person}{Didier Stricker}.} \bibinfo{year}{2021}\natexlab{}.
\newblock \showarticletitle{Synpo-net—Accurate and fast CNN-based 6DoF object pose estimation using synthetic training}.
\newblock \bibinfo{journal}{\emph{Sensors}} \bibinfo{volume}{21}, \bibinfo{number}{1} (\bibinfo{year}{2021}), \bibinfo{pages}{300}.
\newblock


\bibitem[Suzuki et~al\mbox{.}(2020)]%
        {suzuki2020RealitySketch}
\bibfield{author}{\bibinfo{person}{Ryo Suzuki}, \bibinfo{person}{Rubaiat~Habib Kazi}, \bibinfo{person}{Li-yi Wei}, \bibinfo{person}{Stephen DiVerdi}, \bibinfo{person}{Wilmot Li}, {and} \bibinfo{person}{Daniel Leithinger}.} \bibinfo{year}{2020}\natexlab{}.
\newblock \showarticletitle{RealitySketch: Embedding Responsive Graphics and Visualizations in AR through Dynamic Sketching}. In \bibinfo{booktitle}{\emph{Proceedings of the 33rd Annual ACM Symposium on User Interface Software and Technology}} (Virtual Event, USA) \emph{(\bibinfo{series}{UIST '20})}. \bibinfo{publisher}{Association for Computing Machinery}, \bibinfo{address}{New York, NY, USA}, \bibinfo{pages}{166–181}.
\newblock
\showISBNx{9781450375146}
\urldef\tempurl%
\url{https://doi.org/10.1145/3379337.3415892}
\showDOI{\tempurl}


\bibitem[Thalhammer et~al\mbox{.}(2023)]%
        {thalhammer2023challenges}
\bibfield{author}{\bibinfo{person}{Stefan Thalhammer}, \bibinfo{person}{Dominik Bauer}, \bibinfo{person}{Peter H{\"o}nig}, \bibinfo{person}{Jean-Baptiste Weibel}, \bibinfo{person}{Jos{\'e} Garc{\'\i}a-Rodr{\'\i}guez}, {and} \bibinfo{person}{Markus Vincze}.} \bibinfo{year}{2023}\natexlab{}.
\newblock \showarticletitle{Challenges for Monocular 6D Object Pose Estimation in Robotics}.
\newblock \bibinfo{journal}{\emph{arXiv preprint arXiv:2307.12172}} (\bibinfo{year}{2023}).
\newblock


\bibitem[Ulrich et~al\mbox{.}(2022)]%
        {ulrich2022towards}
\bibfield{author}{\bibinfo{person}{Ji\v{r}\'{\i} Ulrich}, \bibinfo{person}{Ahmad Alsayed}, \bibinfo{person}{Farshad Arvin}, {and} \bibinfo{person}{Tom\'{a}\v{s} Krajn\'{\i}k}.} \bibinfo{year}{2022}\natexlab{}.
\newblock \showarticletitle{Towards Fast Fiducial Marker with Full 6 DOF Pose Estimation}. In \bibinfo{booktitle}{\emph{Proceedings of the 37th ACM/SIGAPP Symposium on Applied Computing}} (Virtual Event) \emph{(\bibinfo{series}{SAC '22})}. \bibinfo{publisher}{Association for Computing Machinery}, \bibinfo{address}{New York, NY, USA}, \bibinfo{pages}{723–730}.
\newblock
\showISBNx{9781450387132}
\urldef\tempurl%
\url{https://doi.org/10.1145/3477314.3507043}
\showDOI{\tempurl}


\bibitem[Xiang et~al\mbox{.}(2017)]%
        {xiang2017posecnn}
\bibfield{author}{\bibinfo{person}{Yu Xiang}, \bibinfo{person}{Tanner Schmidt}, \bibinfo{person}{Venkatraman Narayanan}, {and} \bibinfo{person}{Dieter Fox}.} \bibinfo{year}{2017}\natexlab{}.
\newblock \showarticletitle{Posecnn: A convolutional neural network for 6d object pose estimation in cluttered scenes}.
\newblock \bibinfo{journal}{\emph{arXiv preprint arXiv:1711.00199}} (\bibinfo{year}{2017}).
\newblock


\bibitem[Yang et~al\mbox{.}(2022)]%
        {yang2022boosting}
\bibfield{author}{\bibinfo{person}{Hao Yang}, \bibinfo{person}{Chen Shi}, \bibinfo{person}{Yihong Chen}, {and} \bibinfo{person}{Liwei Wang}.} \bibinfo{year}{2022}\natexlab{}.
\newblock \showarticletitle{Boosting 3D Object Detection via Object-Focused Image Fusion}.
\newblock \bibinfo{journal}{\emph{arXiv preprint arXiv:2207.10589}} (\bibinfo{year}{2022}).
\newblock


\bibitem[Zhou et~al\mbox{.}(2020)]%
        {zhou2020gripmarks}
\bibfield{author}{\bibinfo{person}{Qian Zhou}, \bibinfo{person}{Sarah Sykes}, \bibinfo{person}{Sidney Fels}, {and} \bibinfo{person}{Kenrick Kin}.} \bibinfo{year}{2020}\natexlab{}.
\newblock \showarticletitle{Gripmarks: Using Hand Grips to Transform In-Hand Objects into Mixed Reality Input}. In \bibinfo{booktitle}{\emph{Proceedings of the 2020 CHI Conference on Human Factors in Computing Systems}}. \bibinfo{pages}{1--11}.
\newblock


\bibitem[Zhu et~al\mbox{.}(2023)]%
        {zhu2023tracking}
\bibfield{author}{\bibinfo{person}{Jiawen Zhu}, \bibinfo{person}{Zhenyu Chen}, \bibinfo{person}{Zeqi Hao}, \bibinfo{person}{Shijie Chang}, \bibinfo{person}{Lu Zhang}, \bibinfo{person}{Dong Wang}, \bibinfo{person}{Huchuan Lu}, \bibinfo{person}{Bin Luo}, \bibinfo{person}{Jun-Yan He}, \bibinfo{person}{Jin-Peng Lan}, {et~al\mbox{.}}} \bibinfo{year}{2023}\natexlab{}.
\newblock \showarticletitle{Tracking Anything in High Quality}.
\newblock \bibinfo{journal}{\emph{arXiv preprint arXiv:2307.13974}} (\bibinfo{year}{2023}).
\newblock


\bibitem[Zongker et~al\mbox{.}(1999)]%
        {zongker1999environment}
\bibfield{author}{\bibinfo{person}{Douglas~E Zongker}, \bibinfo{person}{Dawn~M Werner}, \bibinfo{person}{Brian Curless}, {and} \bibinfo{person}{David~H Salesin}.} \bibinfo{year}{1999}\natexlab{}.
\newblock \showarticletitle{Environment matting and compositing}. In \bibinfo{booktitle}{\emph{Proceedings of the 26th annual conference on Computer graphics and interactive techniques}}. \bibinfo{pages}{205--214}.
\newblock


\end{thebibliography}

\end{document}